\documentclass[a4paper, 12pt]{article}

\usepackage{amsmath, amssymb, amsfonts}  
\usepackage{cases}
\usepackage{fullpage}

\usepackage{graphicx}    

\usepackage{bm}
\usepackage{bbm}
\usepackage{dsfont}
\usepackage{xcolor}
\usepackage{enumerate}

\usepackage[mathscr]{eucal}
\usepackage{subfigure}
\usepackage{caption}
\usepackage{subcaption}
\usepackage{theorem}

\newtheorem{theorem}{Theorem}

\newtheorem{proposition}{Proposition}
\newtheorem{lemma}{Lemma}

\newtheorem{assumption}{Assumption}

{\theorembodyfont{\upshape}
\newtheorem{remark}{Remark}
\newtheorem{example}{Example}

}

\def\qed{\hfill \vrule height 5pt width 5pt depth 0pt \medskip}
\newcommand{\proof}{\noindent {\bf Proof. }}

\def\diag{{\rm diag}}

\def\rank{{\rm rank}}

\newcommand{\beq}{\begin{equation}}
\newcommand{\eeq}{\end{equation}}
\newcommand{\beqa}{\begin{eqnarray}}
\newcommand{\eeqa}{\end{eqnarray}}
\newcommand{\beqan}{\begin{eqnarray*}}
\newcommand{\eeqan}{\end{eqnarray*}}

\newcommand{\pde}[2]{ \frac{\partial #1}{\partial #2} }

\newcommand{\bite}{\begin{itemize}}
\newcommand{\eite}{\end{itemize}}
\newcommand{\benu}{\begin{enumerate}}
\newcommand{\eenu}{\end{enumerate}}

\definecolor{darkpastelgreen}{rgb}{0.01, 0.75, 0.24}

\begin{document}

\title{\LARGE \bf Multistability of Self-Attention Dynamics in Transformers}

\author{Claudio Altafini\thanks{C. Altafini is with the Division of Automatic Control, Dept. of Electrical Engineering, Link\"oping University, SE-58183, Link\"oping, Sweden. email: {\tt\small claudio.altafini@liu.se}. Work supported in part by the Swedish Research Council (grant n. 2024-04772) and by the ELLIIT framework program at Link\"oping University. }}



\maketitle
%

\begin{abstract}
In machine learning, a self-attention dynamics is a continuous-time multiagent-like model of the attention mechanisms of transformers. In this paper we show that such dynamics is related to a multiagent version of the Oja flow, a dynamical system that computes the principal eigenvector of a matrix corresponding for transformers to the value matrix. 
We classify the equilibria of the ``single-head'' self-attention system into four classes: consensus, bipartite consensus, clustering and polygonal equilibria. Multiple asymptotically stable equilibria from the first three classes often coexist in the self-attention dynamics. 
Interestingly, equilibria from the first two classes are always aligned with the eigenvectors of the value matrix, often but not exclusively with the principal eigenvector.
\end{abstract}


\section{Introduction}

Less than a decade since their introduction \cite{vaswani2017attention}, transformer architectures have become the {\em de facto} standard algorithm for many problems in machine learning and are widely adopted in various fields, such as natural
language processing, computer vision and speech processing \cite{lin2022survey,wen2022transformers,khan2022transformers}.
At the core of a transformer is a so-called self-attention mechanism, a set of operations performed on vectorial representations of the ``tokens'' i.e., elementary units of the objects under analysis (words for large language models (LLM), images patches in computer vision, etc.).
These operations involve three matrices called query ($Q$), key ($K$) and value ($V)$ matrices, two of which, $ Q$ and $ K$, are involved in an inner product, which, exponentiated and normalized by a partition function, yields a softmax function depending on the tokens.
Such softmax function is the celebrated attention mechanism, and expresses how much attention a token $ i$ is giving to another token $j$, relative to the ensemble of all tokens. 
The attention coefficients provide the weight in the weighted sum of the product of the tokens by the value matrix $V$, hence forming a ``self-attention'' mechanism. 
This is the core of a transformer layer, which receives as input the tokens and gives as output ``transformed'' tokens. To avoid a collapsing or exploding token norm due to these operations, the output is then normalized. Other operations (which we do not consider here) are typically present, like for instance multiple such mechanisms act simultaneously to form a ``multi-head'' attention, or the output just described is passed through a feedforward neural network. 
Overall this mechanism constitutes a layer of the transformer: multiple identical layers are concatenated to form what is normally referred to as a transformer.

As is often the case in machine learning, the mathematical understanding of a new approach (``why it works'') lags behind its practical use, and transformer are no exception \cite{van2023large}.
However, given the incredible importance in everyday life that transformer-based applications like LLM are acquiring, investigating and understanding their behavior from a rational perspective appears an important and compelling issue. 

One possible approach to investigating the behavior of transformers was provided recently in a series of papers 
\cite{dutta2021redesigning,sander2022sinkformers,lu2019understanding,geshkovski2023emergence,geshkovski2025mathematical,karagodin2024clustering,wu2024role}.
The basic idea of these papers is to treat the repeated application of identical layers typical of a transformer as the unfolding in time of a dynamical system whose states are the tokens being modified by each transformer layer. 
Rather than dealing with a discrete-time dynamical system (as the setting would immediately suggest), \cite{geshkovski2023emergence,geshkovski2025mathematical} opt for passing to a continuous-time description, which is more amenable to mathematical analysis and easier to characterize. 
The resulting ODE model corresponds to a transformer with an infinite number of layers, which is clearly an idealization (in practical implementations, a transformer may have tens or hundreds of layers).
An interesting perspective that is suggested in \cite{geshkovski2023emergence,geshkovski2025mathematical} is that such ODE can be seen as a multiagent dynamical system, in which each agent (called a ``particle'' in \cite{geshkovski2023emergence,geshkovski2025mathematical}) is a token, and its update law depends on all the other tokens/agents. 
The resulting dynamics is nonlinear due to the attention mechanism, and evolves on a unit sphere because of the normalization operation. 

Multiagent systems on spheres have been studied extensively in the control  community \cite{caponigro2015nonlinear,markdahl2017almost,thunberg2018lifting,zhang2022opinion,zhang2025mixed}, in particular for what concerns colletive phenomena like consensus (all agents converge to the same point in the unit sphere) and more complex, yet related, behaviors like bipartite consensus (when some agents converge to a common point and some other to the antipodal point \cite{caponigro2015nonlinear,zhang2022opinion}) which appear naturally because of the compact nature of the ambient manifold. 
These collective behaviors are highly relevant for transformers: it has in fact been observed repeatedly that transformers indeed tend to be subject to rank-collapse phenomena \cite{dong2021attention,noci2022signal} (also sometimes referred to as token-uniformity or over-smoothing \cite{nguyen2023mitigating,scholkemper2024residual,shi2022revisiting,zhai2023stabilizing,dovonon2024setting}) which appear essentially when the tokens become equal or cluster into groups of equal tokens. 
Indeed in \cite{geshkovski2023emergence,geshkovski2025mathematical} consensus is one of the main behaviors shown to occur for this continuous-time model of self-attention dynamics. A similar result is reported in \cite{abella2024asymptotic} (paper which is closest to ours in terms of mathematical approach).

The scope of this paper is to make a thorough analysis of the asymptotic behavior of the continous-time self-attention dynamics model of \cite{geshkovski2023emergence,geshkovski2025mathematical} using tools from dynamical systems and control. 
In order to do so we establish a connection with another well-know model on the sphere, which, following \cite{oja1982simplified,oja1985stochastic,helmke2012optimization,yan1994global,yoshizawa2001convergence}, we call the Oja flow, but which is also related to the continuous-time Rayleigh quotient flow \cite{helmke2012optimization} and to the continuous-time power method, see eq.~(3) of  \cite{mahony2003continuous}.
This is a much simpler dynamical system whose main feature is that it converges to the principal eigenvector of a matrix which in our setting corresponds to the value matrix $V$. 
In fact, Oja flows are at the basis of algorithms that are used to compute the eigenvectors of a matrix, and have long been used for this scope as an alternative to power methods e.g. in principal component analysis \cite{hornik1992convergence,sanger1989optimal}. 
The Oja flow is insightful but far too simple to use for the self-attention dynamics. 
However a multiagent version of Oja flow, which we develop in the paper, is much more similar, and in fact it corresponds to the self-attention dynamics without the attention coefficients.
Similarly to the Oja flow, the multiagent Oja flow discovers the principal eigenvector of the value matrix $V$, i.e., all agents converge to a consensus equilibrium which is aligned with the principal eigenvector of $V$. 
In addition, the consensus and bipartite consensus points aligned with the other eigenvectors of $V$ are also equilibria, but always unstable. 
It can be shown explicitly that the multiagent Oja flow generically converges to consensus at the principal eigenvector of $V$. 

The self-attention dynamics is obtained inserting an attention matrix in the multiagent Oja flow, and corresponds to replacing a constant average coupling among the agents (equal for all agents) with a weighted average coupling, which is varying from agent to agent and over time.  
Even restricting to time-invariant $ Q$, $ K$ and $V$, and to symmetric $V$, the asymptotic behavior changes significantly w.r.t. the multiagent Oja flow: while consensus at the principal eigenvector still remains a locally asymptotically stable equilibrium point, other locally asymptotically stable equilibria normally emerge, rendering the typical self-attention dynamics multistable.
Most of the new attractors correspond to bipartite consensus equilibria, aligned with the principal eigenvector or with some other eigenvector of $V$, even though sometimes other locally stable equilibria, which we call clustering equilibria, may emerge. 
The name derives from the (numerical) observation that, just like consensus and bipartite consensus, even these extra equilibria are typically in the form of clusters of tokens, i.e., multiple tokens end-up in the same point on the sphere. 
All these equilibria correspond to low-rank attention matrices: rank-1 (uniform) for consensus, rank-2 for a bipartite consensus, and typically low rank also for the clustering equilibria. 
 The complete classification of equilibria of the self-attention dynamics includes also the so-called polygonal equilibria \cite{caponigro2015nonlinear}, which are however always unstable. 
Bipartite consensus and clustering equilibria are not mentioned in papers like \cite{abella2024asymptotic,abellaconsensus}, which focus only on consensus.
While the stability properties of the consensus and bipartite consensus equilibria can be studied analytically through the Lyapunov indirect method, as we do in this paper, for the more complex clustering equilibria an analytical treatment seems still out of reach. 

While convergence towards a consensus-type of equilibrium is known both experimentally and theoretically, the observation that convergence typically occurs towards one of the eigenvectors of the value matrix does not seem to appear in the literature\footnote{With the exception of \cite{abella2024asymptotic}, where asymptotic stability to a consensus aligned with the principal eigenvector of $ V$ is shown, but only for an autoregressive model, i.e., a model with a triangular attention matrix.}. 
Often the consensus or bipartite consensus aligns with the principal eigenvector of $V$, as in a multiagent Oja flow, but alignment with other eigenvectors of $V$ can also occur, in particular with the one corresponding to the least (i.e., most negative) eigenvalue. 
This asymptotic behavior can be considered a form of nonlinear Perron-Frobenius property, and it might provide insight into the interpretation of a transformer, potentially verifiable on real data with pretrained transformer matrices.

\paragraph*{Notation} Boldface letters denote vectors, lower case greek and roman letter scalars, and upper case letters matrices.
The eigenvalues of a matrix $A$ are denoted $ \mu_i [A]$, except for the value matrix $ V$, whose eigenvalues are denoted $ \lambda_i$. 
The inner product is indicated $ \langle \cdot, \cdot \rangle $, while $ A \succeq 0 $ means positive semidefinite (psd). 
The expression $ \bm{x} \parallel \bm{y} $ means $ \bm{y} = \gamma \bm{x} $ for some scalar $ \gamma$, while $ \bm{x} \nparallel \bm{y}$ means no such $ \gamma $ exists. Finally, $ \mathds{1}$ is the vector of all 1.

\section{Model formulation}
Consider $n$ tokens represented as unit length vectors $ \bm{x}_i \in \mathbb{S}^{d-1} \subset \mathbb{R}^d$, $ i=1, \ldots, n $. Denote $ Q,K\in \mathbb{R}^{m\times d}$ the query and key matrices (for simplicity and without loss of generality we hereafter assume $ m=d$)  and $ V \in \mathbb{R}^{d \times d }$ the value matrix. 

Following \cite{geshkovski2025mathematical}, an ODE model for a single-head self-attention mechanism on $n$ tokens can be formulated as follows
\beq
\begin{split}
\dot{\bm{x}}_i & = (I - \bm{x}_i \bm{x}_i^T) \sum_{j=1}^n \frac{e^{\langle Q \bm{x}_i, K \bm{x}_j \rangle } }{\sum_{\ell=1}^n e^{\langle Q \bm{x}_i, K \bm{x}_\ell \rangle } } V \bm{x}_j  \\
& = (I - \bm{x}_i \bm{x}_i^T) V \sum_{j=1}^n A_{ij}(x)  \bm{x}_j, \qquad i =1, \ldots, n
\end{split}
\label{eq:self-att1}
\eeq
where for the terms of \eqref{eq:self-att1} we have the following interpretation:
\bite
\item $ P_i = (I - \bm{x}_i \bm{x}_i^T) $ is the projection onto $ T_{\bm{x}_i}  \mathbb{S}^{d-1}$, the tangent space of $ \mathbb{S}^{d-1} $ at $ \bm{x}_i $. This guarantees that $ \| \bm{x}_i ( t) \|=1 $ for all $ t$, i.e., that the flow of  \eqref{eq:self-att1} evolves on the unit sphere  $ \mathbb{S}^{d-1} $.  In fact, for any $ \bm{y} \in \mathbb{S}^{d-1}$, $ P_i \bm{y} =  (I - \bm{x}_i \bm{x}_i^T) \bm{y} = \bm{y} - \langle \bm{x}_i, \bm{y}\rangle \bm{x}_i $ is always normal to $ \bm{x}_i$, and $ \bm{x}_i^T \dot{\bm{x}}_i =0 $. The projection models the layer normalization present on each layer of the transformer.

\item $ A_{ij}(\bm{x}) = \frac{e^{\langle Q \bm{x}_i, K \bm{x}_j \rangle } }{\sum_{\ell=1}^n e^{\langle Q \bm{x}_i, K \bm{x}_\ell \rangle } } $ is the attention that the token $ \bm{x}_i $ gives to the token $ \bm{x}_j$, computed through a softmax function ($ \bm{x} $ is the stack of $ \bm{x}_1, \ldots, \bm{x}_n $ vectors). $ A_{ij}(\bm{x}) $ is a nonnegative scalar. The attention matrix is then $ A(\bm{x}) = [A_{ij}(\bm{x})]$ and it is a row stochastic matrix, i.e., $ A (\bm{x}) \mathds{1} = \mathds{1}$.
\item In the model \eqref{eq:self-att1} time corresponds to the layer index, hence a self-attention model in continuous-time can be interpreted as a ``continuum of layers''. The asymptotic value of the ODE, $  \bm{x}_i(\infty)$, corresponds to the output of a transformer with an infinite number of layers.
\eite

The model \eqref{eq:self-att1} represents each token $ \bm{x}_i $ as an ``agent'' (it is called a ``particle'' in \cite{geshkovski2025mathematical}) evolving on the sphere $ \mathbb{S}^{d-1}$. 
The total state space is given by $ \bm{x} = \begin{bmatrix} \bm{x}_1^T & \ldots & \bm{x}_n^T \end{bmatrix}^T \in (\mathbb{S}^{d-1})^n$.
Since the flow of \eqref{eq:self-att1} lies on the compact manifold  $  (\mathbb{S}^{d-1})^n$, the vector field is Lipschitz, so forward existence, uniqueness and boundedness for all $t$ follow automatically.

The model \eqref{eq:self-att1} corresponds to an example of collective dynamics on the sphere similar to those investigated in e.g. \cite{caponigro2015nonlinear,markdahl2017almost,zhang2022opinion,zhang2025mixed}: the evolution occurs on a product of unit spheres and it is driven by the interaction with the other agents. 
The difference with these other models of collective dynamics on the sphere is that in \eqref{eq:self-att1} the attention matrix $ A(\bm{x})$ provides the ``interaction graph''. It is typically fully connected and time-varying, since it depends on $ \bm{x}$. 

We are interested in studying the dynamical behavior of \eqref{eq:self-att1}, and in particular in investigating its equilibria and their stability properties. 
To do so, we exploit the fact that the model \eqref{eq:self-att1} has some similarities with the so-called Oja flow, reviewed in next Section, and especially with a multiagent version of Oja flow, investigated in Section~\ref{sec:moja}.


\section{Oja flow}
\label{sec:oja}

In its simplest formulation, the Oja flow \cite{oja1982simplified,oja1985stochastic,helmke2012optimization} is the following dynamical system
\beq
\dot{\bm{x}}  = (I - \bm{x} \bm{x}^T) V \bm{x} , \qquad  \bm{x} \in \mathbb{S}^{d-1} .
\label{eq:oja1}
\eeq
Assume that $ V$ is symmetric of eigenvalues  $ \lambda_1 > \lambda_2 \geq \ldots \geq \lambda_d $, with $ \lambda_1 $ simple, and $ \bm{v}_1, \ldots, \bm{v}_d $ the associated eigenvectors, normalized s.t. $ \| \bm{v}_k \|=1$.
The dynamical behavior of \eqref{eq:oja1} is summarized in the next lemma.
Consider the Rayleigh quotient $ R(\bm{x}) = \frac{\bm{x}^T V \bm{x} }{\| \bm{x} \|^2 }  $, which on the unit sphere reduces to the quadratic form $ R(\bm{x}) = \bm{x}^T V \bm{x}$. 
$R(\bm{x})$ can be used to construct a Lyapunov function for \eqref{eq:oja1}.

\begin{lemma}
\label{lem:oja}
All eigenvectors $ \bm{v}_k $ of $V$ (more precisely, the values $ \pm \bm{v}_k$, $ \| \bm{v}_k\|=1$) are equilibria of \eqref{eq:oja1}.
The function $ W(\bm{x}) = \frac{1}{2} (\lambda_1- R(\bm{x}))$ is a Lyapunov function for \eqref{eq:oja1} and guarantees that \eqref{eq:oja1} converges to the principal eigenvector $ 
\pm \bm{v}_1 $ of $V$ for almost all initial conditions $ \bm{x}(0) \in \mathbb{S}^{d-1}$, while all other $ \pm \bm{v}_k$, $ k=2, \ldots, d$, are unstable.
\end{lemma}

\proof 
From $ \lambda_d = \bm{x}^T \lambda_d \bm{x} \leq R(\bm{x}) \leq \bm{x}^T \lambda_1 \bm{x} = \lambda_1 $, $R(\bm{x})$ is upper bounded by $ \lambda_1 $ on $ \mathbb{S}^{d-1}$, hence $ W(\bm{x})\geq 0 $ and $ W(\bm{x})=0 $ only when $ \bm{x}=\pm \bm{v}_1$, since $ \lambda_1 > \lambda_k $ $k=2, \ldots, d$. 
Differentiating, we have 
\beq
\begin{split}
\dot W(\bm{x}) & = -\bm{x}^T V^2 \bm{x}  + ( \bm{x}^T V \bm{x})^2 \\
& =- \| (I - \bm{x} \bm{x}^T) V \bm{x} \|^2 \leq 0 
\end{split}
\label{eq:oja2}
\eeq
with $ \dot W(\bm{x})=0$  iff $ \bm{x} =\pm \bm{v}_k $ where $ \bm{v}_k $ is an eigenvector of $ V$.  Also, from \eqref{eq:oja2}, $ \dot W(\bm{x})=0$ iff $  (I - \bm{x} \bm{x}^T) V \bm{x}  =0$ i.e., $ V\bm{x} $ is collinear with $ \bm{x}$, which guarantees that the eigenvectors $ \bm{v}_k $ of $V$  (more precisely, on the sphere, the values $ \pm \bm{v}_k $ with $ \| \bm{v}_k \|=1$) are the equilibria of  \eqref{eq:oja1}. 
From LaSalle invariance principle, the only trajectories in the level surfaces of $ W(\bm{x})$ are the eigenvectors $ \bm{v}_k $ of $V$, which guarantees that \eqref{eq:oja2} converges to $ \pm \bm{v}_k $ for some $ k=1, \ldots, d$. 

To show convergence to the principal eigenvector $ \bm{v}_1 $ of $V$ let us consider the linearization of  \eqref{eq:oja1} at $ \bm{v}_k $. Let $ \bm{x} = \bm{v}_k + \bm{u} $ with $ \bm{u}$ a small increment s.t. $ \bm{u}^T \bm{v}_k =0$ (so that the linearization indeed lies in $ T_{\bm{v}_k} \mathbb{S}^{d-1}$, the tangent plane to the unit sphere at $ \bm{v}_k $). Computing the linearization, we get
\beq
\dot{\bm{u}} = (V - \lambda_k I ) \bm{u}.
\label{eq:oja3}
\eeq
Expressing $ \bm{u} $ in the eigenbasis of $V$, $ \bm{u} = \sum_{j=1}^d \eta_j \bm{v}_j $, then for $ j\neq k $, we can project \eqref{eq:oja3} along $ \bm{v}_j $ getting the scalar ODE
$
\dot \eta_j = (\lambda_j - \lambda_k) \eta_j. 
$
If $ k\neq 1 $, then for one of the projections it must be $ j=1$. 
Since $ \lambda_1 $ is a strictly dominant eigenvalue, it is always $ \lambda_1 - \lambda_k>0 $, i.e., each linearization of $ \bm{v}_k \neq \bm{v}_1 $ is unstable. Only when $ \bm{v}_k = \bm{v}_1 $ it is $ \lambda_j - \lambda_1<0 $ for all $j$, meaning that the linearization at $ \bm{v}_1 $ is locally asymptotically stable. Hence \eqref{eq:oja1} converges (almost always) to $ \pm \bm{v}_1 $ i.e. to the principal eigenvector of the matrix $V$. 
\qed

Notice that, while convergence of $ \pm \bm{v}_1 $ is generic, also the other eigenvalues $ \bm{v}_k$, $ k=2, \ldots, d $, have stable submanifolds of various sizes, always strictly smaller than the ambient manifold $ \mathbb{S}^{d-1}$ (and hence of measure 0). 

%


\section{Multiagent Oja flow}
\label{sec:moja}

In this section we propose an extension of the Oja flow in the style of multiagent systems. It corresponds essentially to the self-attention system \eqref{eq:self-att1} without the attention matrix. 
Consider $ n$ vectors $ \bm{x}_i \in \mathbb{S}^{d-1}$ obeying the coupled ODEs:

\beq
\dot{\bm{x}}_i = \frac{1}{n} (I - \bm{x}_i \bm{x}_i^T) \sum_{j=1}^n V \bm{x}_j , \qquad i =1, \ldots, n .
\label{eq:moja}
\eeq
The key difference w.r.t. the Oja flow \eqref{eq:oja1} is that in the right hand side of the ODEs the action of a single agent is replaced by the mean of the $n$ agents $ \bm{m} = \bm{m} ( \bm{x}) =  \frac{1}{n}  \sum_{j=1}^n  \bm{x}_j $.

We make the following assumption which holds throughout the rest of the paper.

\begin{assumption} 
\label{ass:V-sym}
The value matrix $ V\in \mathbb{R}^{d\times d } $ is symmetric, of eigenvalues $ \lambda_1>  \lambda_2 \geq \ldots \geq \lambda_d $ with $ \lambda_1 >0 $ simple and positive.
\end{assumption}
Let $ \bm{v}_1, \ldots, \bm{v}_d $ be the associated eigenvectors, normalized s.t. $ \| \bm{v}_k \|=1$.

\subsection{Equilibria}

Let us begin by expressing the notion of consensus and bipartite consensus for multiagent systems on the sphere that will be used in this paper.
The points $ \bm{x}_1, \ldots, \bm{x}_n \in \mathbb{S}^{d-1} $ are said to be in a {\em consensus state} if $ \bm{x}_i = \bm{x}_j = \bm{v}_k $ $ \forall \, i, j = 1, \ldots, n$ and for some $ k=1, \ldots, d$. 
They are said to be in a {\em bipartite consensus state} if $ \bm{x}_i = \pm \bm{x}_j = \pm \bm{v}_k  $  $ \forall \, i, j = 1, \ldots, n$  and for some $ k=1, \ldots, d$.

\begin{remark}
W.r.t. the literature, \cite{abella2024asymptotic,zhang2022opinion,zhang2025mixed}, we expressly require a consensus or bipartite consensus point to be aligned with one of the eigenvectors of $V$. This choice will be useful when we treat the self-attention dynamics in Section~\ref{sec:self-att}.
\end{remark}

Let $ \bm{y}= V \bm{m} ( \bm{x})  =  \frac{1}{n}V \sum_{j=1}^n  \bm{x}_j \in \mathbb{R}^d$ be the total influence of all agents on agent $i$ (which is the same for all agents).

\begin{lemma}
\label{lem:moja-equil1}
The system \eqref{eq:moja} has the following classes of equilibria:
\benu
\item consensus: $ \bm{x}_i = \bm{v}_k $ $ \forall \, i$, and $ k=1, \ldots, d$;
\item bipartite consensus: $ \bm{x}_i = \pm \bm{v}_k $ $ \forall \, i$ and $ k=1, \ldots, d$;
\item polygonal equilibria: $ \{ {\bm{x}_i \in \mathbb{S}^{d-1}} \; \text{s.t.} \;  V \sum_{j=1}^n  \bm{x}_j =0\}. $
\eenu
\end{lemma}

\proof
The proof follows the reasoning of \cite{caponigro2015nonlinear}. Using $ \bm{y}$, \eqref{eq:moja} becomes 
\beq
\dot{\bm{x}}_i = (I - \bm{x}_i \bm{x}_i^T) \bm{y} = \bm{y}- \bm{x}_i \langle \bm{x}_i, \bm{y} \rangle ,
\label{eq:self_att-axx1}
\eeq
and an equilibrium is a point $ \bm{x} \in (\mathbb{S}^{d-1})^n $ in which $ \bm{y} $ is collinear with $ \bm{x}_i $ $ \forall \, i = 1, \ldots, n$, or in which $ \bm{y}$ vanishes. This can happen in 3 cases:
\benu
\item $ \bm{y} = \gamma_i \bm{x}_i $, for some scalar $ \gamma_i >0$ (consensus);
\item  $ \bm{y} = -\gamma_i \bm{x}_i $, for some scalar $\gamma_i >0$ (bipartite consensus);
\item $ \bm{y} =0$ (polygonal equilibria).
\eenu
In fact, in the first two cases, \eqref{eq:self_att-axx1} becomes $ \dot{\bm{x}}_i = \pm  \gamma_i (\bm{x}_i - \bm{x}_i \bm{x}_i^T \bm{x}_i ) = 0$, since $ \bm{x}_i^T \bm{x}_i =1$. The third case follows trivially from  \eqref{eq:self_att-axx1}.
To show that consensus must be an eigenvector of $V$, observe that at this equilibrium point the total influence $ \bm{y}$ can be written as  $ \bm{y}= V \bm{x}_i $, since $ \bm{x}_i = \bm{x}_j $.
From the expression above, it is also $ \bm{y}= \gamma_i \bm{x}_i $. Putting together these two expressions of $ \bm{y}$: $  V \bm{x}_i =\gamma_i \bm{x}_i $, i.e., $ \bm{x}_i $ is an eigenvector of $V$ and $ \gamma_i $ one of its eigenvalues. The argument for bipartite consensus is identical. 

\qed

For a given $ \bm{v}_k$ there are $ 2^n $ possible consensus or bipartite consensus points.
These are always paired by a global symmetry w.r.t. the origin, i.e., if $ \bm{x}_i = \bm{v}_k $ $ \forall\, i$  is a consensus point, its antipodal point $ \bm{x}_i = - \bm{v}_k $ $ \forall\, i$ is also a consensus point, and similarly for the bipartite consensus equilibria.

The name polygonal equilibria \cite{caponigro2015nonlinear} is due to the observation that $ \bm{z}_j = V \bm{x}_j $ must sum to 0, hence they must correspond to the vertices of a spherical polygon, see also \cite{markdahl2017almost}.
Notice that while the collinear equilibria (consensus and bipartite consensus) are isolated points in $ (\mathbb{S}^{d-1})^n$, polygonal equilibria form instead a set. The set is of zero measure in $ (\mathbb{S}^{d-1})^n$, as it is determined by algebraic constraints. When $ V $ is invertible, the polygonal equilibria are the manifold $ \{ \sum_{j=1}^n \bm{x}_j =0\} \cap (\mathbb{S}^{d-1})^n$.

\subsection{Stability analysis}
The following theorem summarizes the stability and convergence properties of the multiagent Oja system \eqref{eq:moja}.

\begin{theorem}
\label{thm:stab-moja}
For the system \eqref{eq:moja}, under Assumption~\ref{ass:V-sym}, the consensus equilibrium at the principal eigenvector $ \bm{v}_1 $ of $V$ is asymptotically stable, while the other consensus equilibria $ \bm{v}_k $, $ k=2, \ldots, d $, the bipartite consensus equilibria and the polygonal equilibria are all unstable. The trajectories of \eqref{eq:moja} converge to $ \bm{v}_1 $ for almost all initial conditions $ \bm{x}_i(0) \in (\mathbb{S}^{d-1})^n$.
\end{theorem}

To prove this theorem we need a series of preliminary lemmas.
We start by computing the Jacobian linearization of \eqref{eq:moja}. Denote $ f_i(\bm{x}) = \frac{1}{n} (I - \bm{x}_i \bm{x}_i^T) \sum_{j=1}^n V \bm{x}_j  $ and $ f(\bm{x}) =\begin{bmatrix} f_1(\bm{x})^T & \ldots & f_n(\bm{x})^T\end{bmatrix}^T $ the $ (nd) $-dimensional vector field associated to the stacked state vector $ \bm{x}= \begin{bmatrix} \bm{x}_1^T & \ldots & \bm{x}_n ^T \end{bmatrix}^T\in (\mathbb{S}^{d-1})^n$. 

\begin{lemma}
The Jacobian of \eqref{eq:moja}, $ F(\bm{x}) = \pde{f(\bm{x})}{\bm{x}}= \begin{bmatrix}\pde{f_i(\bm{x})}{\bm{x}_h} \end{bmatrix} $ has the following components:
\bite
\item diagonal terms: $$ \!\! \! \! \!\! \! \! \pde{f_i(\bm{x})}{\bm{x}_i} = \frac{1}{n} \Big( (I - \bm{x}_i \bm{x}_i^T) V - \sum_{j=1}^n \left( \bm{x}_i^TV \bm{x}_j I + \bm{x}_i \bm{x}_j^T V \right) \Big), $$
\item off-diagonal terms: $ \pde{f_i(\bm{x})}{\bm{x}_h} = \frac{1}{n} (I - \bm{x}_i \bm{x}_i^T) V $, $ h \neq i $.
\eite
\end{lemma}

\proof
For the diagonal terms direct calculations give
\[
\begin{split}
 \pde{f_i(\bm{x})}{\bm{x}_i} & =\frac{1}{n}  \biggl( \pde{}{\bm{x}_i} V \Big( \sum_{j=1}^n \bm{x}_j \Big) - \pde{}{\bm{x}_i} V \Big(\bm{x}_i \bm{x}_i^T V  \sum_{j=1}^n \bm{x}_j \Big)  \biggl)\\
 & \!\! \! \!  = \frac{1}{n} \biggl( V - \bm{x}_i^T V  \sum_{j=1}^n \bm{x}_j  I - \bm{x}_i \Big( \bm{x}_i^T V + \sum_{j=1}^n \bm{x}_j^T V \Big) \biggl).
 \end{split}
 \]
 For the off-diagonal terms, instead we have
 \[
 \pde{f_i(\bm{x})}{\bm{x}_h}  = \frac{1}{n} (I - \bm{x}_i \bm{x}_i^T) V \pde{}{\bm{x}_h} \sum_{j=1}^n \bm{x}_j .
 \]
 \qed
 
 The linearization can be used to determine the local stability character of the equilibria of \eqref{eq:moja}.
 \begin{lemma}
 \label{lem:lin-stab-moja}
 The consensus point $ \bm{x}_i =\bm{v}_1$ $ \forall\, i$ is locally asymptotically stable for \eqref{eq:moja}, while the remaining consensus equilibria $ \bm{x}_i= \bm{v}_k$ $ \forall \, i $, with $ k= 2, \ldots, d $, are all unstable. At a consensus equilibrium $ \bm{v}_k$, the eigenvalues of $ F(\bm{v}_k) $ are 
 \benu
 \item $ - 2 \lambda_k $ of multiplicity $n$, 
 \item $ \lambda_h - \lambda_k$, for $ h =1, \ldots, k-1, k+1, \ldots, d $,
 \item   $ - \lambda_k $ of multiplicity $ nd -n -d +1 $.
 \eenu
 The bipartite consensus and polygonal equilibria are all unstable.
 \end{lemma}
 
 \proof
 To prove this lemma, we follow the same procedure of Lemma~10 of \cite{zhang2022opinion}.  Remarkably, our system \eqref{eq:moja} and the (different) system in \cite{zhang2022opinion} share the same eigenvectors even though they have different eigenvalues. 
 Notice first that, when computed in an eigenvector $ \bm{v}_k $, the Jacobian of \eqref{eq:moja} can be compactly expressed using tensor products as
 \beq
 \begin{split}
& F(\bm{v}_k)  = \left. \pde{f(\bm{x})}{\bm{x}} \right|_{\bm{x}_i = \bm{v}_k} \\
 & =\frac{1}{n} \mathds{1} \mathds{1}^T \otimes \left( I - \bm{v}_k \bm{v}_k^T  \right)V - I \otimes \left( \bm{v}_k^T V \bm{v}_k I + \bm{v}_k \bm{v}_k^T V \right).
 \end{split}
 \label{eq:JacobF-moja}
 \eeq
 The first term represents a factor present in all entries of $F$, while the second one is present only on the diagonal.
 For $ F(\bm{v}_k) $ there are 3 classes of eigenvectors:
 \benu
 \item  The first class is given by $ \bm{p}^\ell = [ \,\underbrace{0 \, \ldots \, 0}_{\ell-1} \, \bm{v}_k^T \, \underbrace{0 \, \ldots \, 0}_{n- \ell} \,]^T$. There are $ n$ such eigenvectors, and they are obviously all orthogonal to each other. Since 
 \[
  (I- \bm{v}_k \bm{v}_k^T  ) V \bm{v}_k   = \lambda_k (\bm{v}_k - \bm{v}_k \bm{v}_k^T  \bm{v}_k)=0 ,
  \]
  and
  \[
  \begin{split}
 ( \bm{v}_k^T V \bm{v}_k I + \bm{v}_k \bm{v}_k^T V) \bm{v}_k & = \lambda_k (\bm{v}_k^T  \bm{v}_k  \bm{v}_k + \bm{v}_k \bm{v}_k^T  \bm{v}_k) \\
 & =2\lambda_k \bm{v}_k,
 \end{split}
 \]
it is 
\[ 
F(\bm{v}_k ) \bm{p}^\ell = -2 \lambda_k \bm{p}^\ell , \quad \ell =1, \ldots, n ,
\]  
i.e., $ \bm{p}^\ell $ is an eigenvector of $ F(\bm{v}_k) $ of eigenvalue $ - 2 \lambda_k $. 
 
\item The second class of eigenvectors is given by $ \bm{q}^h =\begin{bmatrix} \bm{v}_h^T & \ldots  & \bm{v}_h^T \end{bmatrix}^T$, where $ \bm{v}_h $ is an eigenvector of $V $ associated to $ \lambda_h $ with $ h\neq k$, so that $ \bm{v}_k^T \bm{v}_h =0$. There are $ d-1 $ such  $ \bm{q}^h $ vectors. Computing
\[
 (I- \bm{v}_k \bm{v}_k^T  ) V \bm{v}_h   = \lambda_h (\bm{v}_h - \bm{v}_k \bm{v}_k^T  \bm{v}_h)=\lambda_h \bm{v}_h 
  \]
  \[
 \begin{split}
 ( \bm{v}_k^T V \bm{v}_k I + \bm{v}_k \bm{v}_k^T V) \bm{v}_h & = \lambda_k \bm{v}_k^T  \bm{v}_k  \bm{v}_h + \lambda_h \bm{v}_k \bm{v}_k^T  \bm{v}_h \\
 & =\lambda_k \bm{v}_h ,
 \end{split}
 \]
 hence
 \[ 
 \!\!\!\!
F(\bm{v}_k ) \bm{q}^h =(\lambda_h - \lambda_k) \bm{q}^h ,\;\;  h =1, \ldots, k-1, k+1, \ldots, d .
\]  
\item The remaining $ nd - n - d+1 $ eigenvectors are assembled by considering vectors $ \bm{r} =\begin{bmatrix} (\bm{z}^1)^T & (\bm{z}^2)^T & \ldots  & (\bm{z}^n)^T \end{bmatrix}^T$ s.t. $ \bm{v}_k^T \bm{z}^i =0 $ and $ \bm{v}_h^T \bm{z}^i =0 $ for all $ i =1, \ldots, n $ and all $ h=1, \ldots, k-1, k+1, \ldots, d $. Since the number of such constraints is $ d-1+n$, there exist $ nd- n-d+1 $ such vectors $ \bm{r}$. 
Notice that the $ \bm{z}^i $ can always be chosen so that  $ \sum_{i=1}^n  \bm{z}^i =0 $.
Computing 
\[
  (I- \bm{v}_k \bm{v}_k^T  ) V \bm{z}^i   =V \bm{z}^i -  \lambda_k  \bm{v}_k \bm{v}_k^T \bm{z}^i= V \bm{z}^i 
  \]
  \[
 \begin{split}
 ( \bm{v}_k^T V \bm{v}_k I + \bm{v}_k \bm{v}_k^T V) \bm{z}^i & = \lambda_k \bm{v}_k^T  \bm{v}_k  \bm{z}^i + \lambda_k \bm{v}_k \bm{v}_k^T  \bm{z}_i \\
 & =\lambda_k \bm{z}^i ,
 \end{split}
 \]
 hence
 \[ 
F(\bm{v}_k ) \bm{r}  = \begin{bmatrix} V \sum_i \bm{z}^i - \lambda_k \bm{z}^1 \\ \vdots \\ 
V \sum_i \bm{z}^i - \lambda_k \bm{z}^n  \end{bmatrix} = -\lambda_k \bm{r} .
\]  
Therefore the eigenvalue $ - \lambda_k $ has multiplicity $ nd -n -d +1 $ for $ F(\bm{v}_k)$. 
\eenu
When $ k>1 $, then at least one of the eigenvalues of the second class is $ \lambda_1 -\lambda_k >0$, hence the equilibrium $ \bm{v}_k $ is unstable. When $ k=1$, as by assumption $ \lambda_1>0$ and $ \lambda_1>\lambda_k$ for all $ k=2, \ldots, d$, all eigenvalues of $ F(\bm{v}_1 ) $ are negative, meaning that $ \bm{v}_1 $ is locally asymptotically stable for  \eqref{eq:moja}.

Consider now a bipartite consensus point $ \bm{x}_i = \pm \bm{v}_k$. Assume that $ n_1 $ agents are equal to $ \bm{v}_k $ and $ n_2 =n - n_1$ agent to $ - \bm{v}_k$. Denote $ \mathcal{V}_1 $ and $ \mathcal{V}_2 $ the associated sets of indices.
In the Jacobian matrix $ F(\bm{x}) $ computed at such bipartite consensus we have now the following cases:
\bite
\item if $ i \in \mathcal{V}_1 $ and $ h=i  $:
\[ \!\!\!\!
[F(\bm{v}_k)]_{i h} =
\frac{1}{n}  (I- \bm{v}_k \bm{v}_k^T  ) - \frac{n_1 - n_2 }{n } ( \bm{v}_k^T V \bm{v}_k I + \bm{v}_k \bm{v}_k^T V) 
\]
\item  if $ i \in \mathcal{V}_2 $ and $ h=i$:
\[\!\!\!\!
[F(\bm{v}_k)]_{i h} =
\frac{1}{n}  (I- \bm{v}_k \bm{v}_k^T  ) + \frac{n_1 - n_2 }{n } ( \bm{v}_k^T V \bm{v}_k I + \bm{v}_k \bm{v}_k^T V) 
\]
\item if $ h \neq i  $ 
\[
[F(\bm{v}_k)]_{i h} =
\frac{1}{n}  (I- \bm{v}_k \bm{v}_k^T  ) 
\]
\eite

i.e., some of the diagonal terms switch sign. In compact form, assuming that the first $ n_1 $ indices are in $ \mathcal{V}_1 $ and the remaining $ n_2 $ in $ \mathcal{V}_2 $, 
\[
\begin{split}
 & F(\bm{v}_k) =\frac{1}{n} \mathds{1} \mathds{1}^T \otimes \left( I - \bm{v}_k \bm{v}_k^T  \right)V -  \frac{n_1 - n_2 }{n } \cdot \\
 &  \; \cdot \diag[ \underbrace{1 \; \ldots \; 1}_{n_1 \; \text{times}} \underbrace{-1 \; \ldots \; -1}_{n_2 \; \text{times}} ]  \otimes \left( \bm{v}_k^T V \bm{v}_k I + \bm{v}_k \bm{v}_k^T V \right).
 \end{split}
 \]
Computing eigenvalues in the first of the three classes mentioned above, we have 
\[ 
F(\bm{v}_k ) \bm{p}^\ell = \begin{cases} 
-  \frac{2(n_1 - n_2) }{n } \lambda_k \bm{p}^\ell & \text{if } \; k\in \mathcal{V}_1 \\ 
 \frac{2(n_1 - n_2 )}{n } \lambda_k \bm{p}^\ell & \text{if } \; k\in \mathcal{V}_2.
\end{cases} 
\]  
Regardless of the sign of $ \lambda_k $ and of the cardinality of the $ \mathcal{V}_1 / \mathcal{V}_2 $ partition, the bipartite consensus point is always unstable, since both $ \pm  \frac{2(n_1 - n_2) }{n } \lambda_k  $ are eigenvalues.

Consider now a polygonal equilibrium point $ \bm{x} =\bm{s} =\begin{bmatrix} \bm{s}_1^T & \ldots & \bm{s}_n^T \end{bmatrix}^T \in (\mathbb{S}^{d-1})^n$ where $ \bm{s} $ is s.t. $ \bm{y}= \frac{1}{n} V \sum_{j=1}^n \bm{s}_j  =0$. Let us compute the linearization of \eqref{eq:moja} at $\bm{s}$, obtained perturbing $ \bm{s}$ with a perturbation $ \bm{x}= \bm{s}+ \bm{u} $ belonging to the tangent space $ T_{\bm{s}_i} \mathbb{S}^{d-1}$ of each agent: if $ \bm{u} = [ \bm{u}_1 \ldots \bm{u}_n]$, it is $ \bm{s}_i^T \bm{u}_i =0$, $ i=1, \ldots, n$ and
\[
\begin{split}
\dot{\bm{u}}_i & = \frac{1}{n} \left( I - (\bm{s}_i + \bm{u}_i ) (\bm{s}_i + \bm{u}_i)^T\right) V \sum_{j=1}^n (\bm{s}_j + \bm{u}_j ) \\
& \approx  \frac{1}{n} \left(I - \bm{s}_i\bm{s}_i^T\right) \underbrace{V \sum_{j=1}^n \bm{s}_j}_{=0} + \left(I - \bm{s}_i\bm{s}_i^T\right)V \sum_{j=1}^n \bm{u}_j \\
& \; \; - ( \bm{u}_i \bm{s}_i^T + \bm{s}_i \bm{u}_i^T) \underbrace{V \sum_{j=1}^n \bm{s}_j}_{=0}  + h.o.t.
\end{split}
\]
Recall that $ V$ has always at least one positive eigenvalue $\lambda_1 >0$. We can always choose $ \bm{u}$ s.t. each $ \bm{u}_j $ has a nonzero component in the direction of the associated eigenvector $ \bm{v}_1$: expanding in the eigenbasis of $V$, $ \bm{u}_j = \sum_{k=1}^d \eta^j_k \bm{v}_k $ with $ \eta^j_1\neq 0$. If $ \bm{s}_i = \sum_{k=1}^d \zeta^i_k \bm{v}_k$, and using orthogonality,
\[
\begin{split}
\dot{\bm{u}}_i & = \sum_k \dot \eta^i_k \bm{v}_k \\
& =  \frac{1}{n} \biggl(\Big( I - \sum_k \zeta^i_k \bm{v}_k \sum_\ell \zeta^i_\ell \bm{v}_\ell^T \Big) V \sum_j \sum_k \eta^j_k \bm{v}_k \biggl) \\
& =  \frac{1}{n} \biggl(\Big(  I - \sum_k \zeta^i_k \bm{v}_k \sum_\ell \zeta^i_\ell \bm{v}_\ell^T \Big)  \sum_j \sum_k \eta^j_k \lambda_k \bm{v}_k \biggl) \\
& = \frac{1}{n} \biggl( \sum_{j,k} \eta^j_k \lambda_k \bm{v}_k - \sum_k \zeta^i_k \bm{v}_k \sum_\ell \zeta^i_\ell \lambda_\ell \sum_j \eta^j_\ell  \biggl)
\end{split}
\]
or, projecting along $ \bm{v}_1$, 
\[
\dot  \eta^i_1 = \frac{1}{n} \biggl( \sum_j \eta^j_1 \lambda_1 - \zeta^i_1 \sum_\ell \zeta^i_\ell \lambda_\ell \sum_j \eta^j_\ell \biggl) .
\]
To conclude the argument, it is enough to show that instability occurs along a specific $ \bm{u}$. One such direction is $ \bm{u}$ that perturbs $\bm{s}$ only along the first eigenvector $ \bm{v}_1 $ for each $ i=1, \ldots, n $, i.e., for all $i$, $ \eta^i_1 \neq 0 $ and $ \eta^i_k=0$ for $ k=2, \ldots, d$. In this case in fact we get
\[
\dot  \eta^i_1 = \frac{1}{n} \left( 1- (\zeta^i_1)^2 \right) \lambda_1 \sum_j \eta^j_1 ,
\]
or, in vector form (collecting only the $ \eta^i_1 $ components, $ \bm{\eta}_1 = \begin{bmatrix} \eta^1_1 & \ldots & \eta^n_1 \end{bmatrix}^T$),
\[
\dot{\bm{\eta}}_1 = \frac{\lambda_1}{n} \left( I - \Psi_1^2   \right) \mathds{1} \mathds{1}^T \bm{\eta}_1 
\]
where $ \Psi_1 = \diag(\zeta^1_1, \ldots, \zeta^n_1)$. 
The matrix $ \mathds{1} \mathds{1}^T $ is a rank-1 matrix of eigenvalues $1$ and $0$, while $ |\zeta^i_1 |< 1$ because  $ \bm{s} $ is not an eigenvector of $V$, hence $ I - \Psi_1^2 \succ 0$. Since $ \lambda_1>0$, the matrix $ \frac{\lambda_1}{n} \left( I - \Psi_1^2 \right) \mathds{1} \mathds{1}^T $ has at least one positive eigenvalue, hence $ \bm{s}$ is an unstable equilibrium for \eqref{eq:moja}.
\qed

Finally, the following lemma shows that a Lyapunov stability argument can be set up for \eqref{eq:moja}. 
\begin{lemma}
\label{lem:layp-moja}
The function $ W(\bm{x}) = \frac{1}{2} \left( \lambda_1 - \frac{1}{n} \sum_{j=1}^n \bm{x}_j^T V \sum_{j=1}^n \bm{x}_j  \right)$ is a Lyapunov function for \eqref{eq:moja} and guarantees that \eqref{eq:moja} converges to one of the equilibria determined in Lemma~\ref{lem:moja-equil1}.
\end{lemma}
\proof
Consider $ \bm{m} = \bm{m}(\bm{x}) = \frac{1}{n}  \sum_{j=1}^n  \bm{x}_j $ and $ \bm{y} = V \bm{m}$.
Differentiating $ W(\bm{x}) = \frac{1}{2} \left( \lambda_1 - \frac{1}{n}  \bm{m}^T V \bm{m}  \right)$ gives 
\[
\begin{split}
\dot W(\bm{x}) & = - \frac{1}{n}  \bm{m}^T V\sum_{i=1}^n (I - \bm{x}_i \bm{x}_i^T) V  \bm{m} \\ & = - \frac{n}{n}  \bm{m}^T V V \bm{m} + \frac{1}{n}  \bm{m}^T V \sum_{i=1}^n \bm{x}_i \bm{x}_i^T V  \bm{m} \\ 
& = - \bm{y}^T \bm{y} + \frac{1}{n} \bm{y}^T \sum_{i=1}^n \bm{x}_i \bm{x}_i^T \bm{y} \\ & \leq  - \| \bm{y}\|^2 + \frac{1}{n} \mu_{\max} \left[ \sum_{i=1}^n \bm{x}_i \bm{x}_i^T \right] \| \bm{y}\|^2 \\
& =  \left(- 1 + \frac{n}{n} \right)  \| \bm{y}\|^2=0
\end{split}
\]
where we have used that each psd matrix $ \bm{x}_i \bm{x}_i^T \preceq I $ since it projects onto the direction $ \bm{x}_i$, and hence $ 0 \preceq \sum_{i=1}^n \bm{x}_i \bm{x}_i^T \preceq n I $, from which $ 0 \leq \mu_k \left[\sum_{i=1}^n \bm{x}_i \bm{x}_i^T \right] \leq n$ for all eigenvalues of $ \sum_{i=1}^n \bm{x}_i \bm{x}_i^T$. 

Since the state space is compact, trajectories exist for all $t$ and have limit points. Hence LaSalle's invariance principle applies. In particular, trajectories converge to the largest invariant set contained in 
\[
\begin{split}
\mathcal{L} & = \{ \bm{x} \in (\mathbb{S}^{d-1})^n \;\text{ s. t. }\; \dot W(\bm{x}) =0\} \\ 
& =   \{ \bm{y} \;\text{ s. t. }\;  n \| \bm{y}\|^2 = \sum_{i=1}^n (\bm{x}_i^T \bm{y} )^2 \}.
\end{split}
\]
For the system  \eqref{eq:moja}, $ \bm{x} \in \mathcal{L}$ when $ \bm{y}=0$ or, from the calculations above for $ \dot W$, when equality holds in $  \mu_{\max} \left[\sum_{i=1}^n \bm{x}_i \bm{x}_i^T \right] \leq n$, i.e.,  when all $ \bm{x}_i $ are collinear: $ \bm{x}_i = \pm \bm{v} $ for some $ \bm{v}$. In fact,  in this case, $ \sum_{i=1}^n \bm{x}_i \bm{x}_i^T = n \bm{v} \bm{v}^T $. 
If $ \bm{v} = \bm{v}_k $ for some $ k=1, \ldots, d$, then we have an invariant point. If instead $ \bm{v} \neq \bm{v}_k $, $ k=1, \ldots, d$ (i.e., the consensus point is not an equilibrium), then it is $ \left. \dot{\bm{x}}_i \right|_{\bm{x}_i=\bm{v}} = (I - \bm{v} \bm{v}^T ) V \bm{v} = \left. \dot{\bm{x}}_j \right|_{\bm{x}_j=\bm{v}}  \neq 0$, for all $i, j$, and the dynamics become $n$ identical copies of the Oja flow \eqref{eq:oja1}. From Lemma~\ref{lem:oja}, all these $n$ copies converge to $\pm \bm{v}_k $ for some $k$ (almost always to $ \pm \bm{v}_1 $). 
Summarizing, the largest invariant set in $ \mathcal{L}$ is given by the set of equilibria of \eqref{eq:moja}, hence all trajectories of \eqref{eq:moja} converge to one of the equilibria in $ \mathcal{L}$ computed in Lemma~\ref{lem:moja-equil1}.
\qed

\noindent{\bf Proof of Theorem~\ref{thm:stab-moja}}.
From Lemma~\ref{lem:layp-moja} all trajectories converge to one of the equilibria computed in Lemma~\ref{lem:moja-equil1}.
Lemma~\ref{lem:lin-stab-moja} says that only the consensus point $ \bm{v}_1 $ corresponding to the principal eigenvalue of $V$ is locally asymptotically stable, while all other equilibria are unstable. 
Since all trajectories of \eqref{eq:moja} have a limit point, generically such limit point must be $ \bm{v}_1$. 
\qed

\begin{remark}
Even though all equilibria except one (or one pair, if one counts also the antipodal point) are unstable, they typically have a basin of attraction associated to a stable submanifold. These stable submanifolds are however necessarily of measure 0 in $ (\mathbb{S}^{d-1})^n$, and hence so must be the basins of attraction of the unstable equilibria.
\end{remark}


\section{Self-Attention dynamics}
\label{sec:self-att}

The self-attention system \eqref{eq:self-att1} differs from \eqref{eq:moja} in the fact that in the right hand side of the ODEs, the average of the states of the $n$ agents, $  \frac{1}{n}  \sum_{j=1}^n  \bm{x}_j $, is replaced by a weighted average $ \bm{m} =  \frac{1}{n}  \sum_{j=1}^n  A_{ij}(\bm{x}) \bm{x}_j $, with (state-dependent) weights given by the attention coefficients.

\subsection{Equilibria}

Apart from the three classes of equilibria already obtained for the multiagent Oja flow, in the self-attention dynamics we have an extra class, due to the fact that the total influence on agent $i$ of all other agents $\bm{y}_i = \bm{y}_i (\bm{x}) =  V\sum_{j=1}^n A_{ij}(\bm{x})   \bm{\bm{x}}_j $ now differs from agent to agent.

\begin{lemma}
\label{lem:self-equil1}
The system \eqref{eq:self-att1} has the following classes of equilibria
\benu
\item consensus: $ \bm{x}_i = \bm{v}_k $ $ \forall \, i$, and $ k=1, \ldots, d$;
\item bipartite consensus: $ \bm{x}_i = \pm \bm{v}_k $ $ \forall \, i$ and $ k=1, \ldots, d$;
\item polygonal equilibria: $ \{ \bm{x}_i \in \mathbb{S}^{d-1} \; \text{s.t.} \;  V \sum_{j=1}^n A_{ij}(\bm{x}) \bm{x}_j =0\} $, $ i=1, \ldots, n$;
\item clustering equilibria: $  \{ \bm{x}_i \in \mathbb{S}^{d-1} \; \text{s.t.} \;  \gamma_i \bm{x}_i  = V \sum_{j=1}^n A_{ij}(\bm{x}) \bm{x}_j   \}$ for some scalars $ \gamma_i$, $ i=1, \ldots, n$, $ k=1, \ldots, d$.
\eenu

\end{lemma}
\proof
For the first three cases, the proof is identical to that of Lemma~\ref{lem:moja-equil1}, provided that $ \bm{y} $ is replaced by $\bm{y}_i  =  \sum_{j=1}^n A_{ij}(\bm{x})  V \bm{\bm{x}}_j $. 
Concerning the clustering equilibria, these correspond to $ \bm{y}_i \in \ker(I - \bm{x}_i \bm{x}_i^T) $, or, equivalently, $ \bm{y}_i $ aligned with $ \bm{x}_i $, $ i=1, \ldots, n$ but not necessarily aligned with an eigenvector $ \bm{v}_k$, i.e., $ \bm{y}_i = \gamma_i \bm{x}_i $ for some scalar $ \gamma_i$, but possibly $ \bm{y}_i \nparallel  \bm{v}_k $ $ \forall \, k$.  
\qed

\begin{remark}
Clustering equilibria own their name to the fact that typically multiple agents are found at the same value. In particular we say that $ \bm{x}_1, \ldots, \bm{x}_n$ are at an $ m$-clustering equilibrium point if $ \bm{x}_i \in \{ \bm{w}_1, \ldots, \bm{w}_m \}$ with $ 1 \leq m \leq n$, i.e., if $ \bm{x}_1, \ldots, \bm{x}_n$ cluster at the $m$ vectors $ \bm{w}_1, \ldots, \bm{w}_m$. Notice that some $ \bm{w}_i $ can be eigenvectors of $V$.
\end{remark}

\begin{remark}
A clustering equilibrium can be a consensus, i.e, $ \bm{x}_i  = \bm{x}_j $ for all $ i,j$, but with $ \bm{x}_i $ which is not aligned with any eigenvector of $V$, i.e., $ \bm{x}_i \nparallel  \bm{v}_k $ $ \forall \, k$. We refer to these as 1-clustering, while the ``consensus'' characterization is reserved for the case $ \bm{x}_i \parallel \bm{v}_k $. A 2-clustering instead typically is s.t. $ \bm{x}_i \in \{ \bm{w}_1, \bm{w}_ 2\}$ $ \forall \, i$, with $ \bm{w}_1 \neq - \bm{w}_2 $.
\end{remark}

\begin{proposition}
\label{prop:clustering-eq}
The system \eqref{eq:self-att1} has clustering equilibria iff $ \exists $ scalars $ \gamma_1, \ldots, \gamma_n $ s.t. the matrix $(I_n \otimes I_d  -(\Gamma \otimes I_d) (A(\bm{x}) \otimes I_d)(I_n\otimes V))$ is singular, where $ \Gamma=\diag(\gamma_1, \ldots, \gamma_n )$. 
\end{proposition}
\proof 
In vector form, the clustering equilibrium condition $ \bm{x}_i =  \gamma_i  V \sum_{j=1}^n A_{ij}(\bm{x})  \bm{\bm{x}}_j $ becomes $ \bm{x} = (\Gamma \otimes I_d) (A(\bm{x}) \otimes I_d)(I_n\otimes V)  \bm{x}$, where $ \bm{x}^T = [ \bm{x}_1^T \, \ldots\, \bm{x}_n^T ]$. 
Rewriting it as $  (I_n \otimes I_d -(\Gamma \otimes I_d) (A(\bm{x}) \otimes I_d)(I_n\otimes V)) \bm{x} =0 $, the algebraic equation has nontrivial solutions if and only if the matrix $(I_n \otimes I_d  -(\Gamma \otimes I_d) (A(\bm{x}) \otimes I_d)(I_n\otimes V))$ is singular.
\qed

\begin{remark}
From the proof of Proposition~\ref{prop:clustering-eq}, it follows that if $ \bm{x}_1, \ldots, \bm{x}_n$ form an $m$-clustering equilibrium point, since $ A(\bm{x}) = A(-\bm{x})$, then also the antipodal point  $ -\bm{x}_1, \ldots, -\bm{x}_n $ is an $ m$-clustering equilibrium point. However, `bipartite'' versions of the clustering equilibrium (in which only some $ \bm{x}_i $ flip sign) are typically not equilibria.
\end{remark}

While the attention matrix $ A(\bm{x}) $ is a function of the state even at the equilibrium point (with the exception of a consensus state), its rank is however fixed for various classes of equilibria. 

\begin{lemma}
\label{lem:cons-Aij}
For a consensus state $ \bm{x}_i = \bm{v}_k $ $ \forall \, i$ we have $ A_{ij}(\bm{x}) = \frac{1}{n} $, i.e., the attention matrix $ A(\bm{x}) $ is the rank-1 uniform matrix $ A(\bm{x})  =\frac{1}{n} \mathds{1} \mathds{1}^T $. For a bipartite consensus state $ \bm{x}_i  = \pm \bm{v}_k $ $ \forall \, i$, the attention matrix $A(\bm{x})$ has rank 2.
For an $m$-clustering equilibrium $ \bm{x}$ the rank of $ A(\bm{x})$ is $m$. \end{lemma}
\proof
At a consensus equilibrium $ \bm{x}_i = \bm{v}_k $ for all $ i $, and $ A_{ij}(\bm{x}) $ is composed of all equal terms
\[ 
A_{ij}(\bm{x}) = \frac{e^{  \bm{v}_k^T Q^TK\bm{v}_k  } }{\sum_{\ell=1}^n e^{\bm{v}_k^T Q^TK \bm{v}_k } }= \frac{1}{n}.
\]
At a bipartite consensus point $ \bm{x}_i = \pm \bm{v}_k $, split the $n$ tokens into two sets $ \mathcal{V}_1 $ and $ \mathcal{V}_2 $, $ \mathcal{V}_1\cup \mathcal{V}_2 = \{ 1, \ldots, n \}$,  according to whether $ \bm{x}_i =\bm{v}_k $ or $ \bm{x}_i =- \bm{v}_k$ at equilibrium. Assume that $ n_1 = | \mathcal{V}_1|  $ tokens are equal to $ \bm{v}_k $ and $ n_2 = | \mathcal{V}_2 | $ equal to $ -\bm{v}_k$, with $ n_1 + n_2 =n$.

Four different entries appear in $ A_{ij}(\bm{x})$, two due to the denominator (depending on whether $ \bm{x}_i =\bm{v}_k $ or $ \bm{x}_i =- \bm{v}_k $), and two due to the numerator (depending on whether $ \bm{x}_i$ and $ \bm{x}_j $ have the same sign). 
More specifically, denoting  $ \alpha_1^k = e^{  \bm{v}_k^T Q^TK \bm{v}_k }  $ and $ \alpha_2^k = e^{ - \bm{v}_k^T Q^TK \bm{v}_k } $, then the entries of the attention matrix are 
\[ 
A_{ij}(\bm{x}) 
= \begin{cases} 
\frac{\alpha_1^k}{n_1 \alpha_1^k + n_2 \alpha_2^k} & \text{if}\; i \in \mathcal{V}_1, j \in \mathcal{V}_1 \\ 
\frac{\alpha_2^k}{n_1 \alpha_1^k + n_2 \alpha_2^k}  & \text{if}\; i \in \mathcal{V}_1, j \in \mathcal{V}_2 \\ 
\frac{\alpha_2^k}{n_1 \alpha_2^k + n_2 \alpha_1^k}   & \text{if}\; i \in \mathcal{V}_2, j \in \mathcal{V}_1 \\ 
\frac{\alpha_1^k}{n_1 \alpha_2^k + n_2 \alpha_1^k}   & \text{if}\; i \in \mathcal{V}_2, j \in \mathcal{V}_2  .
   \end{cases}
\]
Letting $ \beta_1^k = n_1 \alpha_1^k + n_2 \alpha_2^k$, and $ \beta_2^k = n_1 \alpha_2^k + n_2 \alpha_1^k$ and assuming w.l.o.g. that the first $ n_1 $ agents are in $ \mathcal{V}_1 $ and the last $ n_2 $ in $ \mathcal{V}_2 $, the attention matrix is 
\beq
A(\bm{x}) = \begin{bmatrix} \frac{1}{\beta_1^k} ( \alpha_1^k & \ldots & \ldots & \alpha_1^k & \alpha_2^k & \ldots & \alpha_2^k) \\
\vdots & & & \vdots & \vdots & & \vdots \\
\frac{1}{\beta_1^k} ( \alpha_1^k & \ldots & \ldots & \alpha_1^k & \alpha_2^k & \ldots & \alpha_2^k) \\
\frac{1}{\beta_2^k} ( \alpha_2^k &  \ldots &  \ldots & \alpha_2^k & \alpha_1^k & \ldots &\alpha_1^k) \\
\vdots & & & \vdots  & \vdots & &  \vdots \\
\frac{1}{\beta_2^k} ( \alpha_2^k & \ldots &  \ldots & \alpha_2^k & \alpha_1^k & \ldots & \alpha_1^k) 
\end{bmatrix} ,
\label{eq:A-bipart}
\eeq
from which it is obvious that the rank must be 2. 

As for an $m$-clustering equilibrium point with vectors $ \bm{w}_1, \ldots, \bm{w}_m$, following a similar procedure it is easy to realize that since $ \bm{w}_i \neq \bm{w}_j $ and $ Q^T K$ is not symmetric,  there are $ m^2 $ different entries in the numerators of $A_{ij}$, and only $m$ in the denominators. Rearranging the entries as in \eqref{eq:A-bipart}, each row of $A$ has exactly $m$ different terms, and a block counting argument leads to $ \rank(A)=m$.
\qed

\subsection{Stability analysis}
In this section we study the stability properties of three of the four classes of equilibria of the self-attention dynamics \eqref{eq:self-att1}. 
Theorem~\ref{thm:stab-main-self} summarizes our main results. 
Before stating it, we need some extra notation.
Consider a bipartite consensus equilibrium associated with the eigenvalue $ \bm{v}_k$ of $V$, and compute $ \alpha_i^k$ and $ \beta_i^k$. 
Denote $ \delta_1^k = \frac{n_1 \alpha_1^k - n_2 \alpha_2^k }{\beta_1^k} $ and $ \delta_2^k =  \frac{n_2 \alpha_1^k - n_1 \alpha_2^k }{\beta_2^k} $. 
 
\begin{theorem}
\label{thm:stab-main-self}
For the self-attention dynamics \eqref{eq:self-att1}, under Assumption~\ref{ass:V-sym}, the consensus equilibrium associated to the principal eigenvector $ \bm{v}_1 $ of $V$ is always asymptotically stable, while the other consensus equilibria $ \bm{v}_k$, $ k=2, \ldots, d $ are all unstable. 
A bipartite consensus equilibrium $ \bm{x}_i = \pm \bm{v}_k$ $ \forall \, i$ is asymptotically stable iff  $ \delta_\ell^k \lambda_k  >0 $, $ \ell =1,2$, 
and the following inequalities are satisfied $ \forall \, j=1, \ldots, k-1, k+1,\ldots,  d $:
\beq
\begin{split}
&  \Big( 1- \frac{\alpha_2^k (n_1^2 + n_2^2 ) }{ 2\alpha_1^k n_1 n_2 }  \Big) \lambda_j - \Big( 1- \frac{(\alpha_2^k)^2}{(\alpha_1^k)^2}  \Big) \lambda_k  <0 \\
& \delta_1^k \delta_2^k  \lambda_k ^2 + \Big( \delta_2^k \frac{n_1 \alpha_1^k}{\beta_1^k}  + \delta_1^k \frac{n_2 \alpha_1^k}{ \beta_2^k} \Big) \lambda_j \lambda_k  \\
& \qquad \qquad  \quad \; \;
+ \frac{n_1 n_2}{\beta_1^k \beta_2^k} \big( (\alpha_2^k)^2 - ( \alpha_1^k)^2 \big) \lambda_j^2  <0 .
\end{split}
\label{eq:cond-gamma-expanded}
\eeq
All polygonal equilibria are unstable.
\end{theorem}

The proof is based only on Lyapunov indirect method.
As for the multiagent Oja flow, it is broken down into various lemmas. 
Letting, as in Section~\ref{sec:moja}, $ f_i(\bm{x})   $ be the right-hand side of \eqref{eq:self-att1} and $ f(\bm{x})  $ its vectorization, we can compute the Jacobian of \eqref{eq:self-att1} as follows.

\begin{lemma}
\label{lem:Jacob-self}
The Jacobian of \eqref{eq:self-att1}, $ F(\bm{x}) = \pde{f(\bm{x})}{\bm{x}}= \begin{bmatrix}\pde{f_i(\bm{x})}{\bm{x}_h} \end{bmatrix} $ has the following components:
\bite
\item diagonal terms: 
\[
\begin{split}
 \pde{f_i(\bm{x})}{\bm{x}_i} & =  (I - \bm{x}_i \bm{x}_i^T) V \Big( \left( I + \bm{x}_i \bm{x}_i^T Q^T K \right) A_{ii}(\bm{x}) \\
&  +  \sum_j \bm{x}_j \big( \bm{x}_j^T K^T Q - \bm{x}_i^TQ^T K A_{ii}(\bm{x}) \\
& - \sum_\ell \bm{x}_\ell^T K^T Q A_{i\ell}(\bm{x}) \big) A_{ij}(\bm{x}) \Big) \\
& - \sum_j \left( \bm{x}_i^T V \bm{x}_j I + \bm{x}_i \bm{x}_j^T V \right) A_{ij} (\bm{x}) ;
\end{split}
\]
\item off-diagonal terms: 
\[
\begin{split}
\pde{f_i(\bm{x})}{\bm{x}_h} & = (I - \bm{x}_i \bm{x}_i^T) V \Big(  I + \bm{x}_k \bm{x}_i^T Q^T K  \\
& \; \; - \sum_j \bm{x}_j \bm{x}_i^T Q^T K A_{ij}(\bm{x}) \Big) A_{ih}(\bm{x}) .
\end{split}
\]
\eite
\end{lemma}

\proof
For simplicity of notation we write $ A_{ij}$ instead of $  A_{ij} (\bm{x})$.
For the diagonal terms we have
\[
\begin{split}
&  \pde{f_i(\bm{x})}{\bm{x}_i}  = V \sum_j \left( \bm{x}_j \pde{A_{ij}}{\bm{x}_i} + A_{ij} \pde{\bm{x}_j}{\bm{x}_i} \right) \\
& \quad  - \bm{x}_i^T V \sum_j A_{ij} \bm{x}_j 
 - \bm{x}_i  \pde{}{\bm{x}_i} \Big( \bm{x}_i^T V \sum_j \bm{x}_j A_{ij} \Big) \\
 & =  V \sum_j  \bm{x}_j \pde{A_{ij} }{\bm{x}_i} + A_{ii} I 
 - \bm{x}_i^T V \sum_j A_{ij} \bm{x}_j I \\
& \quad - \bm{x}_i  \Big(  \sum_j \bm{x}_j^T V  A_{ij} + \bm{x}_i^T V A_{ii} + \bm{x}_i^T V \sum_j \bm{x}_j \pde{A_{ij}  }{\bm{x}_i}  \Big) 
  \end{split}
 \]
 where
 \beq
 \begin{split}
 \pde{A_{ij} }{\bm{x}_i} 
 & = \pde{}{\bm{x}_i} \frac{e^{\langle Q \bm{x}_i, K \bm{x}_j \rangle } }{\sum_{\ell=1}^n e^{\langle Q \bm{x}_i, K \bm{x}_\ell \rangle } }\\
 & = \biggl( \bm{x}_j^T K^T Q e^{\langle Q \bm{x}_i, K \bm{x}_j \rangle }\sum_\ell e^{\langle Q \bm{x}_i, K \bm{x}_\ell \rangle } \\
& \;\;  + \bm{x}_i^T Q^T K e^{\langle Q \bm{x}_i, K \bm{x}_i \rangle } \sum_\ell e^{\langle Q \bm{x}_i, K \bm{x}_\ell \rangle } \delta_{ij}   \\
& - e^{\langle Q \bm{x}_i, K \bm{x}_j \rangle } \Big(  \sum_\ell \bm{x}_\ell^T K^T Q e^{\langle Q \bm{x}_i, K \bm{x}_\ell \rangle } \\
& \;\; + \bm{x}_i^T Q^TK e^{\langle Q \bm{x}_i, K \bm{x}_i \rangle } \Big)  \biggl) / \Big( \sum_\ell \ e^{\langle Q \bm{x}_i, K \bm{x}_\ell \rangle }\Big)^2 \\
& =  \Big( \bm{x}_j^T K^T Q -  \bm{x}_i^T Q^T K A_{ii} - \sum_\ell \bm{x}_\ell^T K^T Q A_{i\ell} \Big) A_{ij} \\
& \;\; +  \bm{x}_i^T Q^T K A_{ii} \delta_{ij}
 \end{split}
 \label{eq:dAij-1}
 \eeq
 with $ \delta_{ij} =1 $ if $ i=j $ and 0 otherwise.
 For the off-diagonal terms, instead we have
 \[
 \begin{split}
 \pde{f_i(\bm{x})}{\bm{x}_h} & =  (I - \bm{x}_i \bm{x}_i^T) V \sum_j \pde{}{\bm{x}_h}  \left( \bm{x}_j A_{ij}\right) \\
 & =  (I - \bm{x}_i \bm{x}_i^T) V \Big( A_{ih} I + \sum_j \bm{x}_j\pde{A_{ij}  }{\bm{x}_h} \Big) ,
 \end{split} 
 \]
 where 
\beq
  \pde{A_{ij} }{\bm{x}_h}   =   \bm{x}_i^T Q^T K \left( A_{ih} \delta_{jh} - A_{ij} A_{ih} \right)  .
 \label{eq:dAij-2}
 \eeq
 \qed
 
Since, from Lemma~\ref{lem:cons-Aij}, in a consensus equilibrium $ A(\bm{\bm{v}_k}) =\frac{1}{n} \mathds{1}\mathds{1}^T$, it is a straightforward computation to show that $ F(\bm{v}_k) $ is still given by \eqref{eq:JacobF-moja}, and hence that $ \bm{v}_1 $ is locally asymptotically stable while $ \bm{v}_k $, $k=2, \ldots, d $, are all unstable.
This is stated in the following lemma.
\begin{lemma}
\label{lem:lin-stab-self1}
The consensus point $ \bm{x}_i =\bm{v}_1$ $ \forall\, i$ is locally asymptotically stable for \eqref{eq:self-att1}, while the remaining consensus equilibria $ \bm{x}_i= \bm{v}_k$ $ \forall \, i $, with $ k= 2, \ldots, d $, are all unstable. At a consensus point $ \bm{v}_k$, the eigenvalues of $ F(\bm{v}_k) $ are those indicated in Lemma~\ref{lem:lin-stab-moja}.
\end{lemma}

The analysis of a bipartite consensus is instead more complicated. 
As in the proof of Lemma~\ref{lem:cons-Aij}, we assume that for the first $n_1 $ agents it is $ \bm{x}_i= \bm{v}_k $, while for the last $ n_2 $ it is $ \bm{x}_i = - \bm{v}_k$. 
In this way the Jacobian has a bipartite structure that reflects that of \eqref{eq:A-bipart}, plus diagonal blocks.

\begin{lemma}
\label{lem:lin-stab-self2}
At a bipartite consensus equilibrium $ \bm{x}_i = \pm  \bm{v}_k$, the Jacobian can be expressed as 
\[
\begin{split}
F(\bm{v}_k) & =  \begin{bmatrix} F_d^1 & \\ &  \ddots \\ & &  F_d^1 \\
& & &  F_d^2 \\ & & & & \ddots \\ & & & & &  F_d^2   \end{bmatrix} \\
& +  \begin{bmatrix} 
F_o^{11} & \ldots &  F_o^{11} &  F_o^{12}   & \ldots &  F_o^{12} \\ 
\vdots & & \vdots & \vdots & & \vdots \\
F_o^{11} & \ldots &  F_o^{11} &  F_o^{12}   & \ldots &  F_o^{12} \\ 
F_o^{21} & \ldots &  F_o^{21} &  F_o^{22}   & \ldots &  F_o^{22} \\ 
\vdots & & \vdots & \vdots & & \vdots \\
F_o^{21} & \ldots &  F_o^{21} &  F_o^{22}   & \ldots &  F_o^{22} \\ 
 \end{bmatrix} ,
 \end{split}
\]
where
\[
\begin{split}
F_o^{11} &=  \frac{\alpha_1^k}{\beta_1^k} (I - \bm{v}_k \bm{v}_k^T) V , \quad i,h\in \mathcal{V}_1 \\
F_o^{12} &=   \frac{\alpha_2^k}{\beta_1^k}(I - \bm{v}_k \bm{v}_k^T) V , \quad i\in \mathcal{V}_1 , \; h \in \mathcal{V}_2  \\
F_o^{21} &=  \frac{\alpha_2^k}{\beta_2^k}(I - \bm{v}_k \bm{v}_k^T) V , \quad i\in \mathcal{V}_2 , \; h \in \mathcal{V}_1  \\
F_o^{22} &=  \frac{\alpha_1^k}{\beta_2^k} (I - \bm{v}_k \bm{v}_k^T) V, \quad i,h\in \mathcal{V}_2 
\\
F_d^1 &= - \delta_1^k ( \bm{v}_k^T V \bm{v}_k I + \bm{v}_k \bm{v}_k^T V ) , \quad i=h\in \mathcal{V}_1 \\
F_d^2 &=- \delta_2^k ( \bm{v}_k^T V \bm{v}_k I + \bm{v}_k \bm{v}_k^T V ) , \quad i=h\in \mathcal{V}_2 .
\end{split}
\]
The eigenvalues of $ F(\bm{v}_k)  $ are
 \benu
 \item $ - 2 \delta_1^k \lambda_k $ of multiplicity $n_1$ and $ - 2 \delta_2^k \lambda_k $ of multiplicity $n_2$;
 \item $ \gamma_{j,\pm}^k =  \frac{1}{2} \Big( a_j^k+ d_j^k \pm \sqrt{ (a_j^k-d_j^k)^2 + 4 c_j^kb_j^k } \Big) $ for $ j=1, \ldots, k-1, k+1,\ldots, d $, where $ a_j^k$, $ b_j^k$, $ c_j^k $ and $ d_j^k$ are given by
 \[
 \begin{split}
 a_j^k & = - \delta_1^k   \lambda_k + \lambda_j \frac{n_1 \alpha_1^k}{\beta_1^k},  \qquad 
 b_j^k = \lambda_j \frac{n_2 \alpha_2^k}{\beta_1^k}   \\
 c_j^k & = \lambda_j \frac{n_1 \alpha_2^k}{\beta_2^k},  \qquad 
 d_j^k  = - \delta_2^k   \lambda_k + \lambda_j \frac{n_2 \alpha_1^k}{\beta_2^k} .
 \end{split}
 \label{eq:a-d} 
 \]
 \item   $ - \delta_1^k \lambda_k $ and $ - \delta_2^k \lambda_k $ of total multiplicity $ nd -n -2d +2 $.
 \eenu
The bipartite consensus equilibrium point is locally asymptotically stable iff $ \delta_\ell^k \lambda_k  >0 $, $ \ell =1,2$, and $  \gamma_{j,\pm}^k<0$ for $ j=1, \ldots, k-1, k+1, \ldots, d$.
\end{lemma}

 \proof
 The formula for the Jacobian at a bipartite consensus point can be obtained from Lemma~\ref{lem:Jacob-self}. After some tedious calculations one gets 

\bite
\item $ i,h\in \mathcal{V}_1  $
\[
F_o^{11} =  (I - \bm{v}_k \bm{v}_k^T) V \left( I + \frac{2 n_2 \alpha_2^k }{\beta_1^k} \bm{v}_k \bm{v}_k ^T Q^T K \right) \frac{\alpha_1^k}{\beta_1^k} ,
\]
\item $ i\in \mathcal{V}_1 , \; h \in \mathcal{V}_2$
\[
F_o^{12} =  (I - \bm{v}_k \bm{v}_k^T) V \left( I - \frac{2 n_1 \alpha_1^k }{\beta_1^k} \bm{v}_k \bm{v}_k ^T Q^T K \right) \frac{\alpha_2^k}{\beta_1^k}, 
\]
\item $ i\in \mathcal{V}_2 , \; h \in \mathcal{V}_1 $
\[
F_o^{21} =  (I - \bm{v}_k \bm{v}_k^T) V \left( I - \frac{2 n_2 \alpha_1^k }{\beta_2^k} \bm{v}_k \bm{v}_k ^T Q^T K \right) \frac{\alpha_2^k}{\beta_2^k},
\]
\item $  i,h\in \mathcal{V}_2 $
\[
F_o^{22} =  (I - \bm{v}_k \bm{v}_k^T) V \left( I + \frac{2 n_1 \alpha_2^k }{\beta_2^k} \bm{v}_k \bm{v}_k ^T Q^T K \right) \frac{\alpha_1^k}{\beta_2^k}, 
\]
\item $  i=h\in \mathcal{V}_1 $
\[
\begin{split}
F_d^1 &=   (I - \bm{v}_k \bm{v}_k^T) V \left( \frac{ 4 n_1 n_2  \alpha_2^k}{\beta_1^k}  \bm{v}_k \bm{v}_k^T   K^T Q  \right) \frac{\alpha_1^k}{\beta_1^k} \\
& - \frac{n_1 \alpha_1^k - n_2 \alpha_2^k }{\beta_1^k} ( \bm{v}_k^T V \bm{v}_k I + \bm{v}_k \bm{v}_k^T V ) ,
\end{split}
\]
\item $  i=h\in \mathcal{V}_2 $
\[
\begin{split}
F_d^2 &=  (I - \bm{v}_k \bm{v}_k^T) V \left( \frac{4 n_1 n_2 \alpha_2^k}{\beta_2^k}  \bm{v}_k \bm{v}_k^T   K^T Q  \right)\frac{\alpha_1^k}{\beta_2^k}  \\
& + \frac{n_1 \alpha_2^k - n_2 \alpha_1^k }{\beta_2^k} ( \bm{v}_k^T V \bm{v}_k I + \bm{v}_k \bm{v}_k^T V ) .
\end{split}
\]
\eite

The expressions in the statement of the lemma follow if one observes that $  (I - \bm{v}_k \bm{v}_k^T) V \bm{v}_k  =0 $.
%
For $ F(\bm{v}_k) $ there are 3 classes of eigenvectors
 \benu
 \item  First class: $ \bm{p}^\ell = [ \,\underbrace{0 \, \ldots \, 0}_{\ell-1} \, \bm{v}_k^T \, \underbrace{0 \, \ldots \, 0}_{n- \ell} \,]^T$. There are $ n$ such eigenvectors, and they are obviously all orthogonal to each other. Since $ (I - \bm{v}_k \bm{v}_k^T) V \bm{v}_k =0 $, it is $  F_o^{ij} \bm{v}_k =0$ for all $ i,j=1,2$, while $  F_d^1 \bm{v}_k = -2 \lambda_k \delta_1^k \bm{v}_k $ and  $  F_d^2 \bm{v}_k = -2 \lambda_k \delta_2^k \bm{v}_k $.
This means that 
 \[ 
F(\bm{v}_k ) \bm{p}^\ell = \begin{cases}
-2 \lambda_k \delta_1^k \bm{p}^\ell ,& \quad \ell =1, \ldots, n_1 \\
- 2 \lambda_k \delta_2^k  \bm{p}^\ell ,& \quad \ell =n_1+1, \ldots, n
\end{cases}
\]  
i.e., $ \bm{p}^\ell $ is an eigenvector of $ F(\bm{v}_k) $, $ \ell =1, \ldots, n$. 
 
\item Second class: Consider the $ d-1$  vectors  $ \bm{q}^h $ s.t. $ \bm{q}^h =[\, \underbrace{ (\bm{w}_1^h)^T \, \ldots  \,  (\bm{w}_1^h)^T}_{n_1 \; \text{times}} \; \underbrace{(\bm{w}_2^h)^T \, \ldots  \, (\bm{w}_2^h)^T}_{n_2 \; \text{times}} ]^T $, with 
$ \bm{w}_1^h = \sum_{j=1}^d \eta_j^{h,1} \bm{v}_j $ and $ \bm{w}_2^h = \sum_{j=1}^d \eta_j^{h,2} \bm{v}_j $.
We have $ F_o^{ij} \bm{w}_\ell^h = \frac{\alpha_j}{\beta_i} \sum_{m\neq k } \lambda_m \eta_m^{h,\ell} \bm{v}_m $ for $ i, j, \ell = 1,2$ 
and $ F_d^i \bm{w}_i^h = -\delta_i^k \lambda_k (\bm{w}_i^h + \eta^{h,i}_k \bm{v}_k ) $, $ i = 1,2 $.
The $ \bm{q}^h $ are obtained solving the algebraic equation  $ F  \bm{q}^h = \gamma \bm{q}^h $ where also the eigenvalue $ \gamma $ is an unknown.
Expanding we obtain a block of $ n_1 $ identical equations
\[
\begin{split}
 & - \delta_1^k   \lambda_k (\bm{w}_1^h + \eta^{h,1}_k \bm{v}_k ) + \sum_{j\neq k} \lambda_j \left( \frac{n_1 \alpha_1^k}{\beta_1^k} \eta^{h,1}_j  \right. \\
 & \;\; \left. + \frac{n_2 \alpha_2^k}{\beta_1^k} \eta^{h,2}_j \right) \bm{v}_j  = \gamma \sum_{j=1}^d \eta_j^{h,1} \bm{v}_j 
 \end{split}
 \]
 and another of $ n_2 $ identical equations
 \[
 \begin{split}
 &- \delta_2^k   \lambda_k (\bm{w}_2^h + \eta^{h,2}_k \bm{v}_k ) + \sum_{j\neq k} \lambda_j \left( \frac{n_1 \alpha_2^k}{\beta_2^k} \eta^{h,1}_j \right. \\
 & \quad \left.  + \frac{n_2 \alpha_1^k}{\beta_2^k} \eta^{h,2}_j \right) \bm{v}_j = \gamma \sum_{j=1}^d \eta_j^{h,2} \bm{v}_j
 \end{split}
 \]
 whose solution provides both the desired eigenvector $ \bm{w}_i^h $ and the associated eigenvalues $ \gamma$. 
As can be seen projecting along $ \bm{v}_k $, these equations have solution only if $ \eta^{h,\ell}_k =0 $ i.e., if both $ \bm{w}_1^h $ and $ \bm{w}_2^h $ are orthogonal to $ \bm{v}_k$: $ \bm{v}_k^T  \bm{w}_\ell^h=0$. (similarly to the second class of eigenvectors in the multiagent Oja flow case, see Lemma~\ref{lem:lin-stab-moja}).
Projecting these equations along the eigenvector $ \bm{v}_j $ and rearranging
\[
\begin{split}
\Big(  \overbrace{ \!- \delta_1^k   \lambda_k + \lambda_j \frac{n_1 \alpha_1^k}{\beta_1^k}}^{=a_j^k} - \gamma \Big) \eta^{h,1}_j \! + \! \Big( \overbrace{\!\lambda_j \frac{n_2 \alpha_2^k}{\beta_1^k} }^{=b_j^k} \Big)  \eta^{h,2}_j  & = 0 \\
\Big( \underbrace{\!  \lambda_j \frac{n_1 \alpha_2^k}{\beta_2^k}}_{=c_j^k} \Big) \eta^{h,1}_j  \!+\! \Big( \underbrace{\! - \delta_2^k   \lambda_k + \lambda_j \frac{n_2 \alpha_1^k}{\beta_2^k}  }_{=d_j^k} - \gamma  \Big)  \eta^{h,2}_j  & = 0.
\end{split}
\]
Notice that $ a_j^k$, $b_j^k $, $ c_j^k$ and $ d_j^k $ are independent of the index $h$, hence, to avoid repeated identical algebraic equations, we can take $ h=j$ and drop one index in the $ \eta_j^{h,i}$ variables: $ \eta_j^{h,i} = \eta_j^i $, obtaining
\[
\begin{split}
(a_j^k-\gamma) \eta^1_j  +b_j^k  \eta^2_j & =0 \\
c_j^k \eta^1_j  + (d_j^k-\gamma)   \eta^2_j  & =0 ,
\end{split}
\]
which leads to the formula for the eigenvalues $ \gamma_{j, \pm}^k = \frac{1}{2} \Big( a_j^k+ d_j^k \pm \sqrt{ (a_j^k-d_j^k)^2 + 4 c_j^kb_j^k } \Big) $. Since $ b_j^kc_j^k>0 $, the two solutions are always real. The relationship between the components of the two vectors $ \bm{w}_\ell^j $ is then $  \eta^2_j = -\frac{a_j^k- \gamma_{j, \pm}^k}{b_j^k} \eta^1_j$.
Notice that for each $ \bm{q}^j $ there are two eigenvalues $ \gamma_{j, \pm}^k$, for a total of $ 2 (d-1)$ eigenvalues in this class.

\item 
The third class contains the $ nd - n - 2d+2 $ remaining eigenvectors subdivided into two groups: $ \bm{r}_1 = \begin{bmatrix}(\bm{z}^1)^T & \ldots &  (\bm{z}^{n_1})^T & 0 & \ldots & 0 \end{bmatrix}^T$ and $ \bm{r}_2 = \begin{bmatrix} 0 & \ldots & 0  & (\bm{z}^{n_1+1})^T & \ldots &  (\bm{z}^n)^T \end{bmatrix}^T$ where $ \bm{z}^i $ s.t. $ \bm{v}_k^T \bm{z}^i =0 $ and $ (\bm{w}_\ell^j)^T \bm{z}^i =0 $ for all $ i =1, \ldots, n $, $ j=1, \ldots, k-1, k+1, \ldots, d $ and $ \ell =1,2$. 
The $ \bm{z}^i $ are chosen so that  $\sum_{i=1}^{n_1}  \bm{z}^i  =0 $ and $ \sum_{i=n_1+1}^{n}  \bm{z}^i =0 $.
Computing: $ F_o^{ij} z^\ell = \frac{\alpha_j}{\beta_i} V \bm{z}^\ell $ for all $ i,j=1,2$, and $F_d^i \bm{z}^\ell = - \delta_i \lambda_k \bm{z}^\ell $, $ i=1,2$.
Hence 
\[
F(\bm{v}_k ) \bm{r}_1
= \begin{bmatrix}  
 \frac{\alpha_1^k}{\beta_1^k}  \sum_{j=1}^{n_1} V \bm{z}^j - \delta_1^k \lambda_k \bm{z}^1 \\ 
\vdots \\ 
 \frac{\alpha_1^k}{\beta_1^k}  \sum_{j=1}^{n_1} V \bm{z}^j - \delta_1^k \lambda_k \bm{z}^{n_1} \\ 
\frac{\alpha_2^k }{\beta_2^k}  \sum_{j=1}^{n_1} V \bm{z}^j   \\ 
\vdots \\
 \frac{\alpha_2^k }{\beta_2^k}  \sum_{j=1}^{n_1} V \bm{z}^j 
  \end{bmatrix} = - \delta_1^k  \lambda_k \bm{r}_1
\]  
and, similarly, $ F(\bm{v}_k ) \bm{r}_2 = - \delta_2^k  \lambda_k \bm{r}_2$. 
\eenu

Concerning stability, notice that the eigenvalues of $ F(\bm{v}_k) $ in the first and third class depend exclusively from the sign of $ \lambda_k$ and $ \delta_\ell^k$, while eigenvalues  depending on the difference $ \lambda_j - \lambda_k $ no longer appear directly, even though $ \lambda_j $ and $ \lambda_k $ enter into the complicated expressions of the second class, which varies with the cardinality of the splitting $ \mathcal{V}_1/\mathcal{V}_2$ in the  bipartite consensus equilibria associated to $ \bm{v}_k$. 
What can be concluded straightforwardly is that a bipartite consensus is stable iff $ \delta_\ell^k \lambda_k >0$, $ \ell =1,2$ and $ \gamma_{j,\pm}^k $ are all in the left half plane. 
\qed

\begin{lemma}
\label{lem:polyg-self}
The polygonal equilibria are all unstable for \eqref{eq:self-att1}.
\end{lemma}
\proof
The idea of the proof is similar to that used in the multiagent Oja flow. 
A polygonal equilibrium point $ \bm{x} =\bm{s} =\begin{bmatrix} \bm{s}_1 & \ldots & \bm{s}_n \end{bmatrix} \in (\mathbb{S}^{d-1})^n$ is s.t. $ V  \sum_{j=1}^n A_{ij}(\bm{s}) \bm{s}_j  =0$. Let us compute the linearization of \eqref{eq:self-att1} at $\bm{s}$, obtained perturbing $ \bm{s}$ with a perturbation $ \bm{u} = [ \bm{u}_1 \ldots \bm{u}_n] $ with  $  \bm{u}_i  \in T_{\bm{s}_i} \mathbb{S}^{d-1} $. Retaining only the first order terms:
\[
\begin{split}
\dot{\bm{u}}_i & = \frac{1}{n} ( \bm{u}_i \bm{s}_i^T + \bm{s}_i \bm{u}_i^T)  \underbrace{V \sum_{j=1}^n  A_{ij}(\bm{s}) \bm{s}_j}_{=0} \\
& + \left(I - \bm{s}_i\bm{s}_i^T\right) V \sum_j \sum_h (\bm{z}_h^{ij})^T \bm{u}_h \bm{s}
_j \\
& + \left(I - \bm{s}_i\bm{s}_i^T\right) V  \sum_{j=1}^n  A_{ij}(\bm{s}) \bm{u}_j + h.o.t.
 \\
\end{split}
\]
where the vectors $ (\bm{z}_h^{ij})^T  = \pde{A_{ij}(\bm{s})}{\bm{u}_h} $ are computed in \eqref{eq:dAij-1} and \eqref{eq:dAij-2}.
Denote $ \xi^{ij} = \sum_h (\bm{z}_h^{ij})^T \bm{u}_h$ and observe that $ \xi^{ij} $ is a sum of bilinear forms $ \bm{s}_\ell B_k \bm{u}_h $ for some matrices $ B_k $ (see \eqref{eq:dAij-1} and \eqref{eq:dAij-2} for their specific expressions). 
From the matrix Cauchy-Schwartz inequality, for each of these bilinear forms it holds $ - \| B_k\|_2 \leq \bm{s}_\ell B_k \bm{u}_h \leq \| B_k \|_2 $, where $ \|B_k \|_2 $ is  independent of $ \bm{u}_h$.

Expanding in a basis of eigenvectors of $ V$: $ \bm{s}_i = \sum_{k=1}^d \zeta^i_k \bm{v}_k$ and $ \bm{u}_j = \sum_{k=1}^d \eta^j_k \bm{v}_k $, we get
 \[
\begin{split}
\dot{\bm{u}}_i & = \sum_k \dot \eta^i_k \bm{v}_k 
=  \biggl(\Big( I - \sum_k \zeta^i_k \bm{v}_k \sum_\ell \zeta^i_\ell \bm{v}_\ell^T \Big) V \cdot  \\
& \;\; \cdot \sum_j \Big( \xi^{ij} \sum_\ell \zeta^\ell_k \bm{v}_k + A_{ij}(\bm{s}) \sum_k \eta^j_k \bm{v}_k \Big) \biggl) .
\end{split}
\]
Assuming that the perturbation $ \bm{u}$ is aligned with $ \bm{v}_1$, i.e., for all $i$, $ \eta^i_1 \neq 0 $ and $ \eta^i_k=0$ for $ k=2, \ldots, d$, then, projecting along $ \bm{v}_1$, yields
\[
\dot  \eta^i_1 =\left( 1- (\zeta^i_1)^2 \right) \lambda_1 \Big( \sum_j \xi^{ij}\zeta^j_1 +  \sum_j A_{ij}(\bm{s}) \eta^j_1 \Big) .
\]
Denoting  $ \bm{\eta}_1 = \begin{bmatrix} \eta^1_1 & \ldots & \eta^n_1 \end{bmatrix}^T$
and $  \bm{\zeta}_1 = \begin{bmatrix} \zeta^1_1 & \ldots & \zeta^n_1 \end{bmatrix}^T$ the collection of the $ \eta_1^i $ and $ \zeta_1^i $ components of all agents, the previous ODEs can be expressed in vector form as
\beq
\dot{\bm{\eta}}_1 = \lambda_1 \left( I- \Psi_1^2 \right) \left( \Xi \bm{\zeta}_1 + A(\bm{s} ) \bm{\eta}_1 \right)
\label{eq:ode-eta-self}
\eeq
where $ \Psi_1 = \diag(\bm{\zeta}_1)$, $ \Xi =[\xi^{ij}]$ is a matrix with lower and upper bounds independent of $ \bm{u}$, and $ \bm{\zeta}_1 $ is fixed. $ A(\bm{s} ) $ is a row stochastic matrix,  $ \rho(A(\bm{s} )) =1 $ and $ A(\bm{s} ) \mathds{1} = \mathds{1}$. 
Since $ |\zeta^i_1 |< 1$ because  $ \bm{s} $ is not an eigenvector of $V$ and $ \lambda_1 >0$, the system \eqref{eq:ode-eta-self} diverges when $ \bm{\eta}_1 = \epsilon \mathds{1}$ for some scalar $ \epsilon$. Hence the polygonal equilibrium $ \bm{s}$ is unstable. 
\qed

\noindent {\bf Proof of Theorem~\ref{thm:stab-main-self}.}
The proof is the direct combination of Lemmas~\ref{lem:lin-stab-self1},~\ref{lem:lin-stab-self2} and~\ref{lem:polyg-self}, with the only observation that in the second class of eigenvalues of the bipartite consensus case of Lemma~\ref{lem:lin-stab-self2}, the formula for the eigenvalues can be written equivalently as $ \gamma_{j, \pm}^k = \frac{1}{2} \Big( a_j^k+ d_j^k \pm \sqrt{ (a_j^k+d_j^k)^2 - 4 a_j^k d_j^k + 4 c_j^kb_j^k } \Big) $, from which the condition $ \gamma_{j, \pm}^k <0$ becomes $ a_j^k + d_j^k <0 $ and $ c_j^k b_j^k > a_j^k d_j^k $ $ \forall \, j=1, \ldots, k-1, k+1,\ldots,  d $. From these, after some calculations, one gets \eqref{eq:cond-gamma-expanded}.
\qed

\begin{remark}
Since $ A(\bm{x}) = A(- \bm{x}) $, the global symmetry between any equilibrium point $ \bm{x}$ and its antipodal point $ - \bm{x}$ is preserved, hence $ \bm{x} $ and $ - \bm{x}$ have the same stability properties. 
\end{remark}

The characterization of Theorem~\ref{thm:stab-main-self} is weaker than that of Theorem~\ref{thm:stab-moja} in several aspects, because the behavior of \eqref{eq:self-att1} is significantly more complex than that of \eqref{eq:moja}. 
In particular, while in \eqref{eq:moja} there is an almost globally asymptotically stable attractor, the hallmark of \eqref{eq:self-att1} is its multistability. 
We list now some of the limitations of Theorem~\ref{thm:stab-main-self}:
\bite
\item The stability character of the bipartite consensus equilibria cannot be determined a priori, as cannot be the $ \bm{v}_k$ to which they align.
\item The stability character of the clustering equilibria could not be verified analytically. In particular obtaining an explicit expression of the Jacobian linearization and of its eigenvalues seems out of reach for now. What we can see in simulations is that clustering equilibria (typically of low $m$) can be locally asymptotically stable, but seem to be rarer than the bipartite consensus equilibria.
\item The total number of coexisting attractors cannot be determined a priori.
\item No Lyapunov-like function with globally nonincreasing derivative could be found for \eqref{eq:self-att1}.
\eite

We also notice that for this multistable system computing the basin of attraction of the various equilibria seems a difficult problem. The simple case $ V=I$ treated  in \cite{abella2024asymptotic} in terms of hemispheres does not obey to Assumption~\ref{ass:V-sym}.

\section{Numerical examples}

In this section we investigate some small-scale examples numerically, to get some insight into the behavior of the system \eqref{eq:self-att1}.

\begin{example}
\label{ex:3dgeneral-1}
For $ d=3 $ and $ n=10$, and for randomly chosen $ Q, K$ and $ V$, examples of trajectories are given in Fig.~\ref{fig:3Dexamples2}. In panels (a) and (b) the agents converge to consensus equilibria, in panel (c) to a bipartite consensus equilibrium and in panel (d) to a 3-clustering equilibrium.
In the bipartite consensus case, the stable equilibrium point is aligned with $ \bm{v}_1$.
Notice how sometimes the transient of a trajectory can be long and irregular (see example of panel (b)), which reminds somehow of the idea of a transient ``metastability''  mentioned in  \cite{geshkovski2025mathematical}.
\begin{figure*}[ht]
     \centering    
      \subfigure[consensus]{
        \includegraphics[trim=2cm 0.5cm 2cm 1cm, clip=true,width=0.4\textwidth]{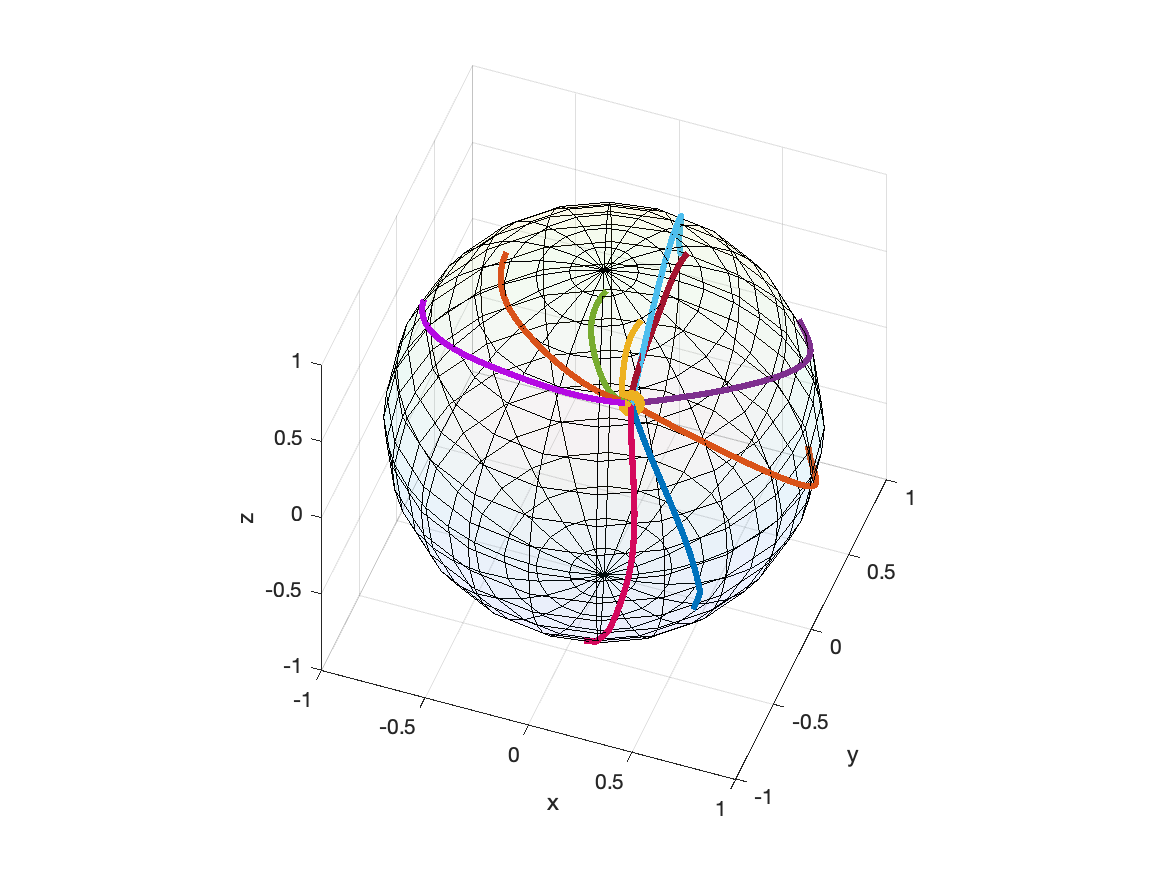}}
      \subfigure[consensus, longer transient]{
        \includegraphics[trim=2cm 0.5cm 2cm 1cm, clip=true,width=0.34\textwidth]{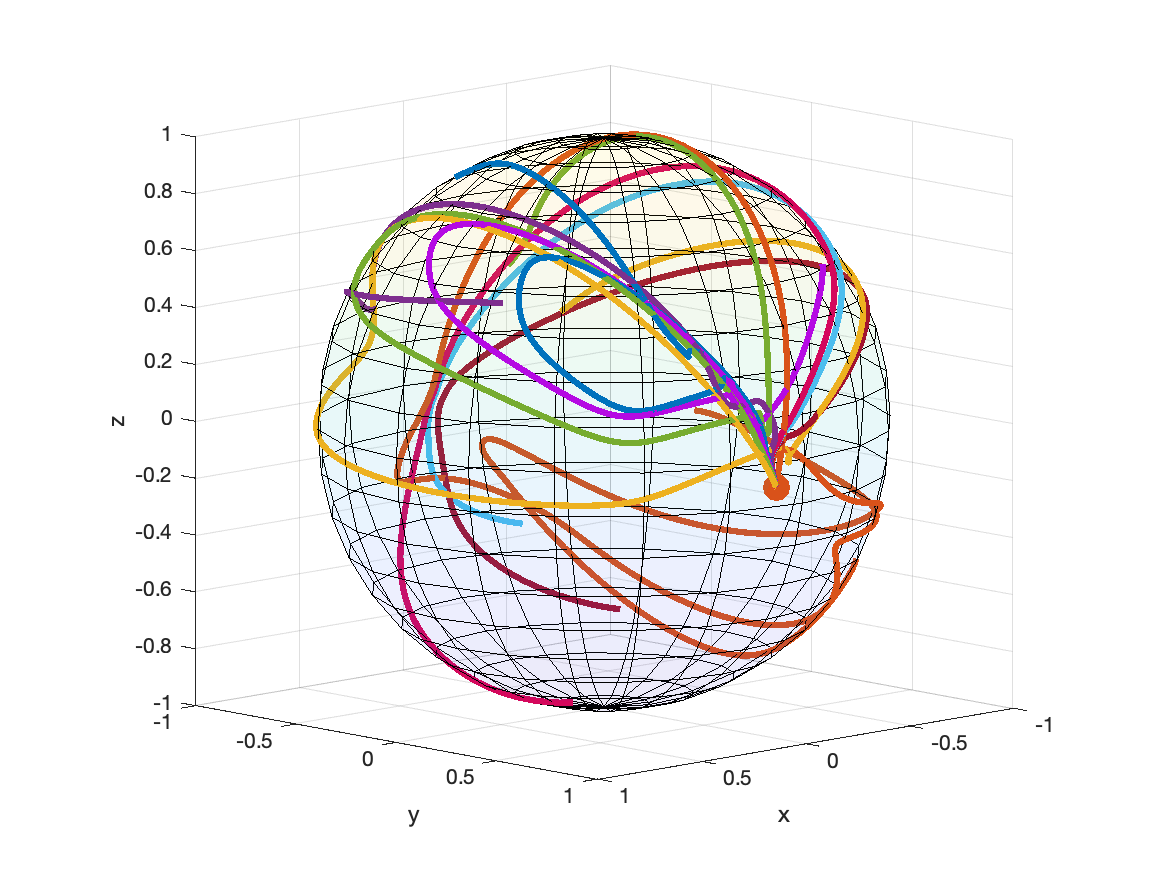}}
   \subfigure[bipartite consensus]{
        \includegraphics[trim=2cm 0.3cm 2cm 1cm, clip=true,width=0.4\textwidth]{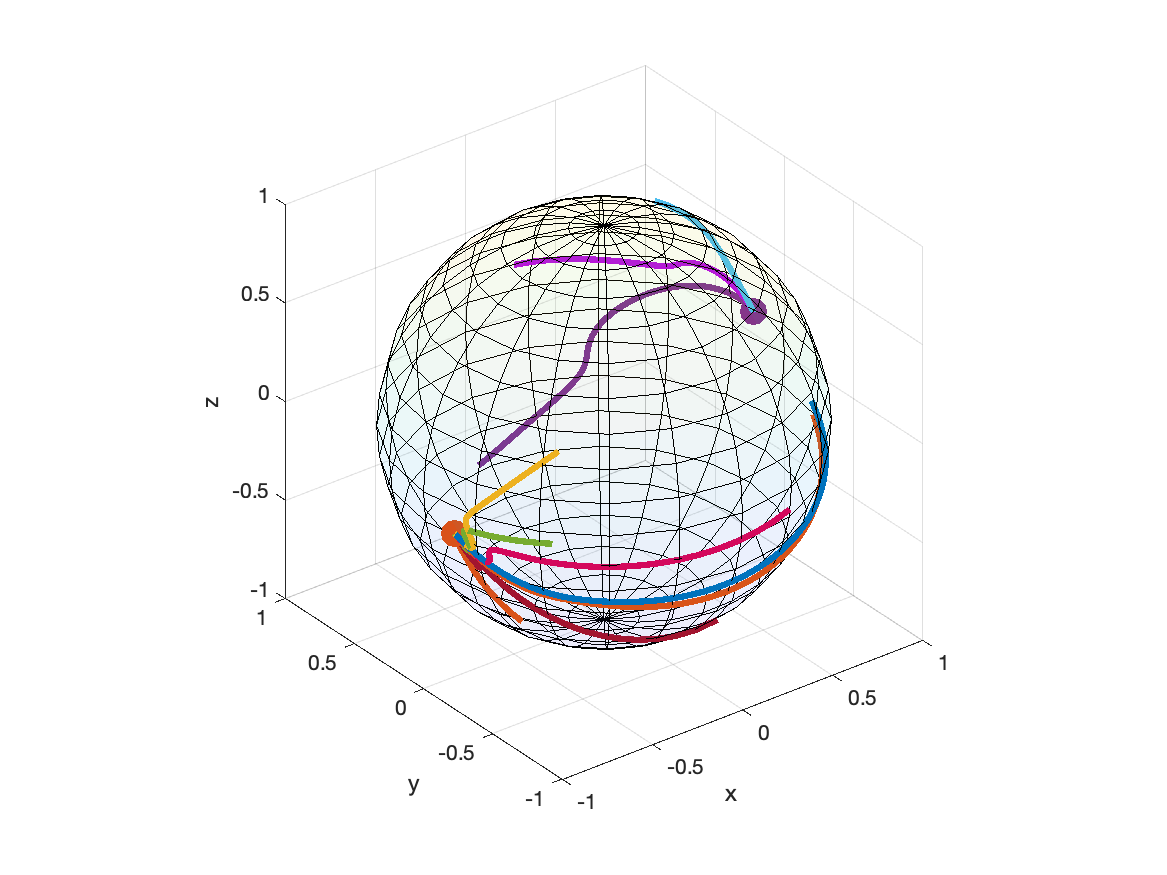}}
  \subfigure[3-clustering]{
      \includegraphics[trim=2cm 0.5cm 2cm 1cm, clip=true,width=0.4\textwidth]{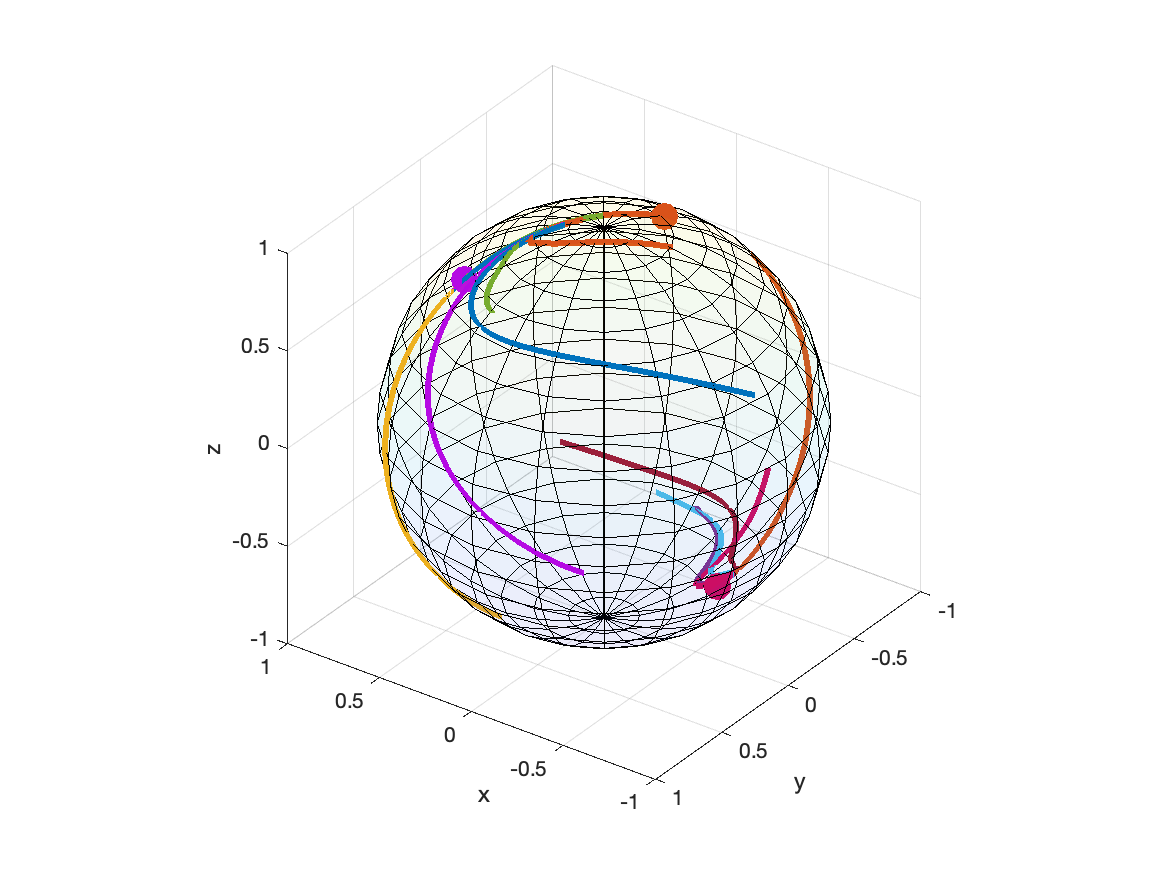}}
        \caption{Example~\ref{ex:3dgeneral-1}, with $ d=3$ and $ n=10$. The solid dot is the endpoint of a trajectory.}
        \label{fig:3Dexamples2}
\end{figure*}
\end{example}

\begin{example}
\label{ex:equilibria-1}
We consider now an example with $ d=20$ and $ n=100$. 
We generate randomly 50 instance of the matrices $ Q$, $ K$ and $V$, and for each triplet we perform 100 simulations, all leading to an equilibrium point. 
These equilibria are classified into consensus, bipartite consensus (specifying also to which eigenvector $ \bm{v}_k $ they align with) and clustering, specifying also $m$, the number of clusters. 
The resulting values are shown in Fig.~\ref{fig:equilibria-1}.
Recall that a 1-clustering is a consensus equilibrium not aligned with any eigenvector $ \bm{v}_k$. 
A 2-clustering equilibrium is instead in general not a bipartite consensus.
In more than 50\% of the instances multistability appears.

\begin{figure}[ht]
     \centering    
      \includegraphics[trim=0cm 0cm 0cm 0cm, clip=true,width=0.6\textwidth]{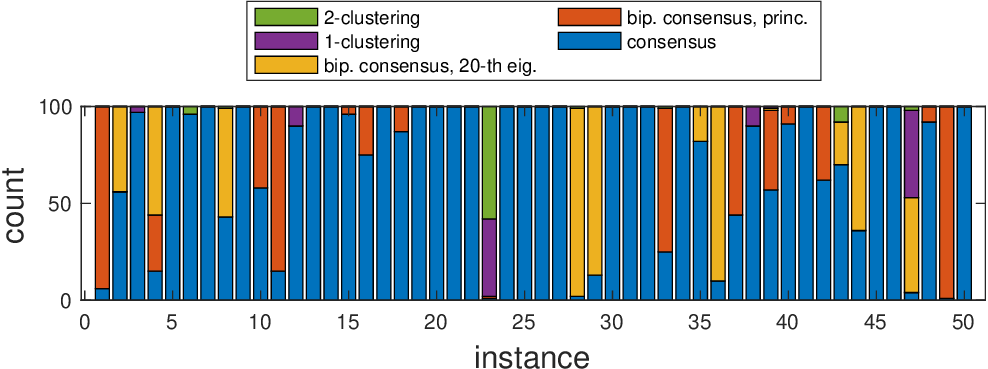}
        \caption{Example~\ref{ex:equilibria-1}, numerical classification of stable equilibria.}
        \label{fig:equilibria-1}
\end{figure}
\end{example}

\begin{example}
\label{ex:equilibria-2}
In this example we aim to check the local stability of all consensus and bipartite consensus equilibria associated to all eigenvectors $ \bm{v}_k$, $ k=1, \ldots, d$. Due to their explosion in number ($ 2^n$) this can be done exhaustively only for small scale systems. Here we choose $ d=4 $ and $ n=10$. 
Fig.~\ref{fig:equilibria-2}(a) shows that in 100 instances we tested, out of $ d 2^n =4096$ such equilibria, in some cases nearly half can be stable. In some other cases, instead, only the consensus aligned with $ \bm{v}_1 $ (and its antipodal point) are instead stable, depending on the choice of $ V$, $ Q$ and $K$. 
Interestingly, the stable bipartite consensus equilibria are always aligned with the principal eigenvector $ \bm{v}_1 $ or with the least (i.e., most negative) eigenvector $ \bm{v}_4 $. 
Whenever the latter case occurs, it is always $ | \lambda_4| > \lambda_1 $. 
Convergence to $ \bm{v}_1 $ and $ \bm{v}_4 $ can coexist in a system.

In Fig.~\ref{fig:equilibria-2}(b) we consider instead a larger system, $ d=10$ and $n=100$, and for each $ \bm{v}_k $ we sample 100 bipartite consensus equilibria for each eigenvector $ \bm{v}_k$. The situation is very similar: out of a total of 1000 equilibria, a fraction varying between 1 and 200 is stable, and convergence to bipartite consensus aligned with $ \bm{v}_1 $ dominates, followed by $ \bm{v}_{10}$. None of the other eigenvectors has any stable equilibrium.

\begin{figure}[ht]
     \centering    
      \subfigure[]{
      \includegraphics[trim=0cm 0cm 0cm 0cm, clip=true,width=0.6\textwidth]{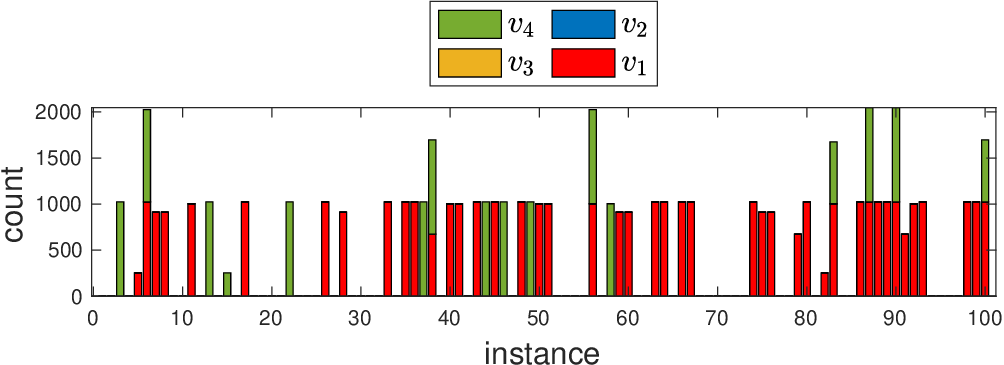}}
      \subfigure[]{
      \includegraphics[trim=0cm 0cm 0cm 0cm, clip=true,width=0.6\textwidth]{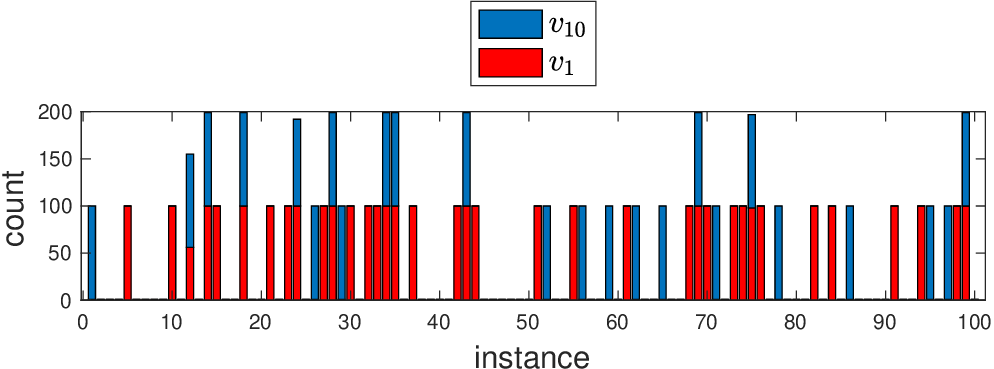}}
        \caption{Example~\ref{ex:equilibria-2}. (a): exhaustive count of all stable consensus + bipartite consensus equilibria; (b): random sampling stable bipartite consensus equilibria.}
        \label{fig:equilibria-2}
\end{figure}
\end{example}

%
%

\section{Extensions of the model}

In formulating the model \eqref{eq:self-att1} we made a series of simplifying assumptions, which are now commented upon.
\bite
\item {\em $V$ symmetric and with a simple, positive principal eigenvalue}. Numerically we see that this assumption can be relaxed as long as the principal eigenvalue of $V$ remains real and simple. When a complex conjugate pair becomes the principal eigenvalue of $V$, then the self-attention dynamics may converge to a stable limit cycle. 
It remains to understand whether bipartite consensus or clustering equilibria are still present in this case, and what is their stability character.

\item {\em A scaling factor $ \beta$ is disregarded} in the inner product leading to the attention matrix. 
This scaling factor is sometimes defined as $ \beta =\frac{1}{\sqrt{d}}$, but in principle it can be interpreted as an inverse temperature. Including it, the attention matrix becomes 
\beq
A_{ij}(\bm{x}) = \frac{ e^{\beta \langle Q \bm{x}_i, K \bm{x}_j \rangle } }{\sum_{\ell=1}^n e^{\beta \langle Q \bm{x}_i, K \bm{x}_\ell \rangle } } .
\label{eq:A-beta}
\eeq
One can study the behavior of \eqref{eq:self-att1} in the various possible regimes of $ \beta$, see \cite{geshkovski2025mathematical}. As stated in next proposition, when $ \beta\to0 $ we recover the multiagent Oja flow \eqref{eq:moja}.
\begin{proposition}
The self-attention model \eqref{eq:self-att1} with the attention coefficients \eqref{eq:A-beta} collapses into the multiagent Oja flow \eqref{eq:moja} when $ \beta\to 0$. 
\end{proposition}
\proof
Just observe that when $ \beta \to 0$, $ e^{\beta \langle Q \bm{x}_i, K \bm{x}_j \rangle } \to 1 $, hence $ A_{ij}(\bm{x}) \to \frac{1}{n} $, regardless of $ \bm{x}$.
\qed

\item {\em The model \eqref{eq:self-att1} uses a ``single-head'' attention mechanism}, instead of a ``multi-head'' attention.  
A multihead self-attention dynamics looks like
\[
\dot x_i = (I - x_i x_i^T)  \sum_{h=1}^H V_h \sum_{j=1}^n A_{h,ij}(x)  x_j .
\]
It is trivial to show that consensus is still an asymptotically stable equilibrium point, with the single principal eigenvector of $V$ replaced by a combination of principal eigenvectors of all $ V_h$.
The analysis of the other equilibria and of their stability is instead more complex and will be discussed in another venue.

\item {\em Continuous-time instead of discrete-time}.  A similar analysis can be carried out in discrete-time. In fact, the discrete-time model can be considered an Euler discretization of the continuous-time model \cite{geshkovski2025mathematical,abella2024asymptotic}.
\item {\em Time-invariant $ Q$, $ K$ and $ V$}, instead of time-varying. In the time-varying case, the analysis becomes more challenging, because asymptotic stability must be shown in a uniform sense. See \cite{abella2024asymptotic} for some progress in this direction.

\item {\em No feedforward neural network}. This is impossible to include in the continuous-time model. See again \cite{abella2024asymptotic} for comments on what happens when it is added in discrete-time.

\eite

\section{Conclusion}
For the self-attention dynamical model of a transformer, in this paper we carried out a thorough analysis of the landscape of equilibria and investigated their stability properties. 
A feature that emerges is that multistability often occurs, associated typically, but not exclusively, to consensus (or consensus-like equilibria, like bipartite consensus). 
Another feature is that these stable consensus-like equilibria are aligned with the eigenvectors of the value matrix $V$, typically with the principal eigenvector, but sometimes also with other eigenvectors. 
If this property is confirmed also in more realistic models, it suggests that each layer of a transformer may act by tilting a token vector towards one of the eigenvectors of the value matrix, a property that we plan to verify experimentally in the near future. 

\section{Acknowledgments}
This work was done while the author was on sabbatical at MIT. The hospitality of Ali Jadbabaie and IDSS is gratefully acknowledged.


\end{document}